\PassOptionsToPackage{table}{xcolor}
\documentclass{article}
\usepackage{iclr2027_conference,times}
\iclrfinalcopy

\usepackage{amsmath,amsfonts,bm}

\def\eqref#1{equation~\ref{#1}}

\def\1{\bm{1}}

\DeclareMathAlphabet{\mathsfit}{\encodingdefault}{\sfdefault}{m}{sl}
\SetMathAlphabet{\mathsfit}{bold}{\encodingdefault}{\sfdefault}{bx}{n}

\usepackage{marvosym}
\usepackage{graphicx}
\usepackage{booktabs}
\usepackage{makecell}
\usepackage{xspace}
\usepackage{multirow}
\usepackage{tabularx}
\usepackage{pifont}
\usepackage{subcaption}
\usepackage{amssymb}
\usepackage{xcolor}
\definecolor{navyblue}{HTML}{0071BC}
\usepackage[hidelinks]{hyperref}

\newcommand{\name}{DynThinker\xspace}
\newcommand{\sota}{\textbf}
\newcommand{\cmark}{\ding{51}}

\newcommand{\xmarkg}{\textcolor{lightgray}{\ding{55}}}

\title{DynThinker: Unifying Static 3D and Dynamic 4D Spatial Reasoning}

\author{
Qihang Cao$^{*1}$, Haoyuan Li$^{*2}$, Zihan Guo$^{2}$, Sensen Gao$^{1}$, Zihui Gao$^{1}$, Huixin Zhang$^3$,\\[5pt]
\textbf{Fan He}$^{3}$, \textbf{Jianhua Han}$^{3}$, \textbf{Hang Xu}$^{3}$, \textbf{Xiaodan Liang}$^{2,3,4}$, \textbf{Jia-Wang Bian}$^{1}$\textsuperscript{\Letter}\\[5pt]
$^1$Nanyang Technological University,
$^2$Shenzhen campus of Sun Yat-sen University, \\[5pt]
$^3$Yinwang Intelligent Technology Co. Ltd, 
$^4$MBZUAI\\[5pt]
}
\vspace{-3mm}

\begin{document}

\renewcommand{\thefootnote}{$\ast$}
\footnotetext{Equal contribution.}

\renewcommand{\thefootnote}{\Letter}
\footnotetext{Corresponding author.}

\maketitle
\lhead{Preprint}

\begin{center}
    \centering
    \captionsetup{type=figure}
    \begin{center}
        \includegraphics[width=\textwidth]{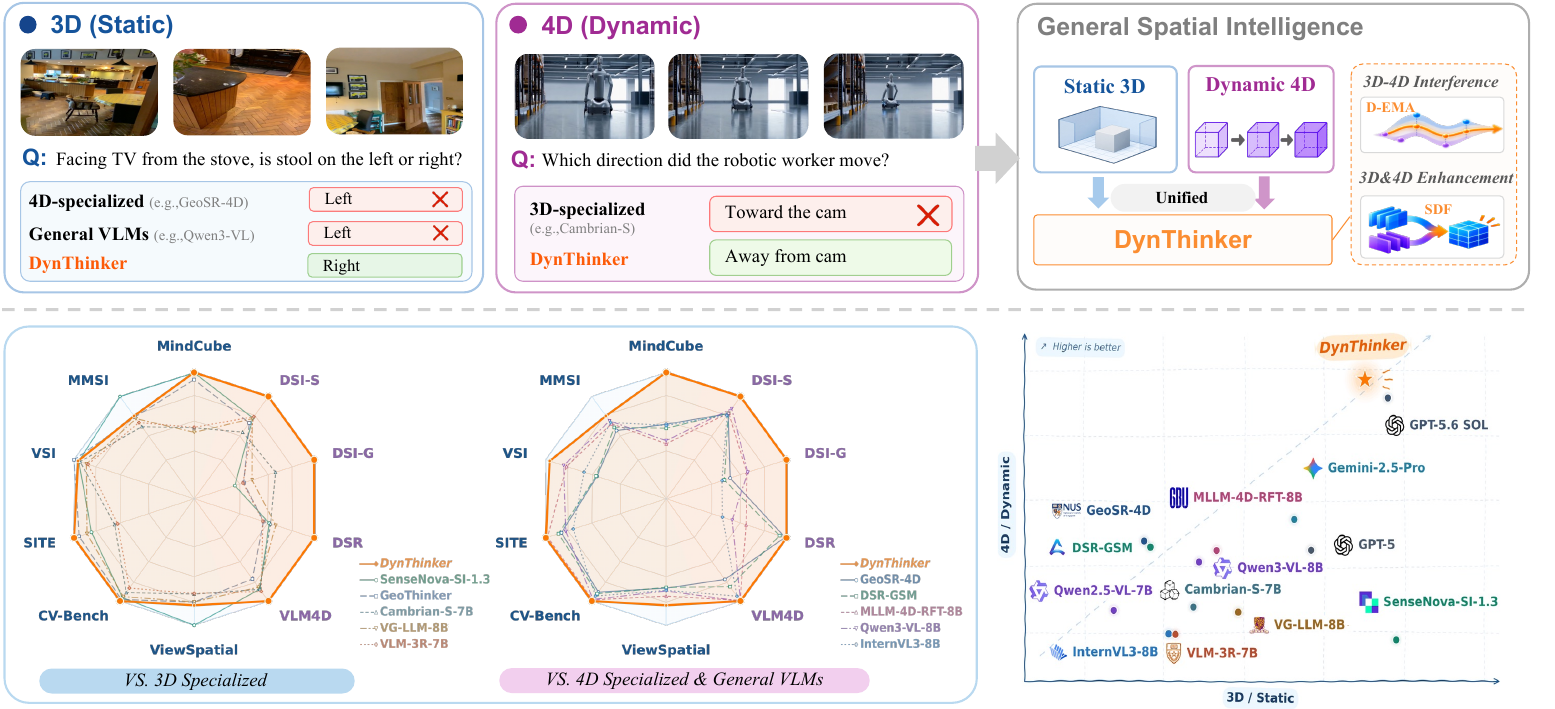}
        
    \end{center}
 \caption{
 \textbf{From Spatial--Dynamic Interference to Joint Improvement.} 
 \textbf{Top}: General spatial intelligence requires both static 3D understanding and dynamic 4D reasoning.
 Existing methods often improve one capability at the expense of the other, leading to spatial--dynamic interference. 
 \name jointly strengthens both capabilities within a unified model.
 \textbf{Bottom}: The radar plots compare \name with 3D-specialized, 4D-specialized, and general VLMs across static and dynamic benchmarks. 
 The 3D--4D capability plane further highlights its strong and balanced performance across both dimensions, moving toward unified spatial reasoning.
 }
    \label{fig:teaser}
\end{center}

\begin{abstract}

Achieving general spatial intelligence requires a unified understanding of both static 3D and dynamic 4D.
However, we find that existing models struggle to excel at both: 3D post-training can weaken dynamic understanding, while 4D-specialized models often lack strong static 3D reasoning.
Our analysis reveals that this spatial--dynamic interference is highly non-uniform across parameter groups, with the balance between static and dynamic capabilities particularly sensitive to changes in late-layer MLPs.
To address this interference, we propose \name, a unified framework for jointly strengthening static and dynamic spatial reasoning. 
Decoupled Exponential Moving Average (D-EMA) applies configurable group-wise averaging to constrain drift in interference-sensitive pretrained parameters while preserving flexibility for capability enhancement.
Spatial-Dynamic Fusion (SDF) complements D-EMA by selectively integrating spatial and temporal priors from a dynamic geometry encoder into intermediate VLM representations under semantic guidance. 
Across ten metrics spanning nine static 3D and dynamic 4D benchmarks, \name achieves the highest average score among all evaluated models. 
Notably, \name achieves 56.0 on ReVSI, 67.2 on DSR-Bench, and 51.2 and 40.1 under the sample-wise and group-wise evaluation protocols of DSI-Bench, respectively, setting new state-of-the-art results among the compared open-source models across all four evaluations within a unified model. 
\end{abstract}
\section{Introduction}
\label{sec:intro}
\vspace{-3mm}

Spatial intelligence has emerged as a central capability for Multimodal Large Language Models (MLLMs), 
enabling them to understand and reason about both static 3D structure and dynamic 4D environments, supporting applications ranging from scene understanding~\citep{yang2025thinking, cao2026vggt(caoyang)} to embodied interaction~\citep{lin2025evo,li2025spatialforcing,yang2026visual}.
Recent post-training methods have substantially improved static 3D reasoning~\citep{tong2024cambrian(cambrian-1),zhang2025flatland(spar),yin2025spatial(mindcube),li2025viewspatial}, while parallel efforts have extended spatial reasoning to dynamic 4D environments~\citep{zhou2025vlm4d,zhou2025DSR-GSM,zhang2025dsi,li2025sti}.
Yet these directions remain largely specialized: 3D-oriented models often lack dynamic understanding, whereas 4D-oriented models remain limited in static 3D reasoning. This gap motivates a unified model capable of strong spatial reasoning across both static and dynamic environments.

However, unified spatial reasoning is challenging, as improving one capability can come at the expense of the other. We find that 3D-focused spatial post-training can interfere with pretrained dynamic competence: improvements in static 3D reasoning are accompanied by substantial degradation on dynamic 4D benchmarks~\citep{zhou2025vlm4d}. This trade-off leads to imbalanced capability profiles shown in Figure~\ref{fig:teaser}, revealing spatial--dynamic interference that hinders joint capability improvement.
Existing capability-preservation strategies provide only partial remedies.
Data replay requires auxiliary data and careful mixing, regularization can limit adaptation, and parameter-efficient methods such as LoRA~\citep{hu2022lora} may restrict update capacity. The \textit{central challenge} is therefore to preserve existing dynamic competence while allowing sufficient flexibility to strengthen static 3D reasoning, without relying on replay data or additional regularization objectives.

To better understand this interference, we analyze 3D-specialized VLMs through checkpoint interpolation and parameter-group restoration. 
We find that the interference is highly non-uniform across parameter groups, with the balance between static and dynamic capabilities particularly sensitive to changes in late-layer MLPs, as shown in Figure~\ref{fig:motivation}.
Restoring these parameters to their pretrained values substantially recovers dynamic performance, while partially reversing the gains in static 3D reasoning. This opposing response indicates that the tension between static 3D enhancement and dynamic capability preservation is particularly pronounced in late-layer MLPs.

Motivated by these observations, we propose \name, a unified spatial reasoning framework that addresses spatial--dynamic interference through complementary designs.
First, \textbf{Decoupled Exponential Moving Average (D-EMA)} mitigates interference through configurable group-wise temporal averaging, allowing stronger consolidation for interference-sensitive parameters while preserving flexibility for capability enhancement. The decay is adjusted according to the number of EMA updates, ensuring consistency across training schedules.
Second, \textbf{Spatial-Dynamic Fusion (SDF)} complements parameter preservation with joint spatial--dynamic perception by selectively integrating spatial and temporal priors from a dynamic geometry encoder into intermediate VLM representations. Guided by the VLM's semantic context, SDF adaptively modulates contribution of each prior stream, enabling model to capture both static geometric structure and temporal evolution.

Extensive experiments demonstrate strong and balanced performance across static 3D and dynamic 4D reasoning. Across ten metrics, \name achieves the highest average score of 60.5, surpassing the proprietary frontier model GPT-5.6 Sol at 59.5, while also attaining state-of-the-art results among the compared open-source models on ReVSI, DSR-Bench, and DSI-Bench. Ablation studies further confirm the complementary roles of D-EMA and SDF in mitigating spatial--dynamic interference while strengthening spatial reasoning, showing that spatial specialization need not come at the expense of dynamic capabilities. 
Our contributions can be summarized as follows:

\vspace{-2mm}
\begin{itemize}

    \item \textbf{Diagnosing Spatial--Dynamic Interference.} 
    We identify a spatial--dynamic interference in which 3D-focused post-training strengthens static reasoning while degrading dynamic competence inherited from pretraining, posing a key challenge to unified spatial reasoning.

    \vspace{-1mm}
    \item \textbf{A Unified Framework for Spatial Reasoning.} 
    To overcome this interference and bridge static 3D and dynamic 4D reasoning, we propose \name, a unified framework that strengthens both capabilities within a single VLM. D-EMA mitigates interference during adaptation, while SDF adaptively integrates spatial and temporal priors for enhancement.

    \vspace{-1mm}
    \item \textbf{Strong and Scalable 3D--4D Performance.} 
    \name achieves strong and balanced performance across ten static 3D and dynamic 4D metrics, attaining state-of-the-art results among compared open-source models on ReVSI, DSR-Bench, and DSI-Bench. 

\end{itemize}

\section{Related Work}
\label{sec:related_work}
\vspace{-1mm}

\subsection{Static and Dynamic Spatial Intelligence}
\vspace{-2mm}

Progress toward real-world spatial intelligence requires vision–language models (VLMs) to reason about both static spatial structure and dynamic changes over time.
Early work explores image-based spatial reasoning, with SpatialVLM learning qualitative and metric spatial relations from automatically generated image--question pairs~\citep{chen2024spatialvlm}.
Beyond individual images, benchmarks such as VSI-Bench and MindCube require models to integrate observations across multiple views and reason about 3D scene layout, spatial relationships, and metric properties~\citep{yang2025thinking,yin2025spatial(mindcube),yang2025mmsi}.
Recent work has extended spatial intelligence toward dynamic 4D settings, with benchmarks spanning complementary data sources and reasoning targets that together motivate unified reasoning across static 3D and dynamic 4D.
VLM4D combines real and synthetic videos to assess motion, viewpoint changes, and temporal continuity~\citep{zhou2025vlm4d}, while DSI-Bench focuses on disentangling observer motion from object motion~\citep{zhang2025dsi}. DSR-Bench instead builds on real-world videos to evaluate evolving 3D relations and multi-object interactions~\citep{zhou2025DSR-GSM}, and MLLM4D-Bench uses scenarios derived from stereo videos to probe spatial and temporal reasoning~\citep{yin2026mllm4d}. 
Building on these efforts, we study how to jointly strengthen static and dynamic spatial capabilities during post-training.

\vspace{-1mm}
\subsection{Geometry Integration in Vision--Language Models}
\vspace{-2mm}

To equip VLMs with stronger 3D and 4D reasoning, methods augment semantic visual features with explicit geometric cues~\citep{wang2026feed}.
Geometry encoders include VGGT, which recovers 3D scene structure from multiple views~\citep{wang2025vggt}, and V-DPM, which extends geometric reconstruction to dynamic scenes with 3D motion estimation~\citep{sucar2026v(VDPM)}. Existing methods incorporate geometric knowledge into VLMs through input-level feature fusion~\citep{zheng2025learning(vgllm)}, distillation~\citep{li2025spatialforcing}, and intermediate-layer integration~\citep{li2026thinking}.
For static and dynamic spatial reasoning, DSR-GSM uses $\pi^3$~\citep{wang2026pi3} to capture both scene geometry and motion, together with stacked Q-Formers to select question-relevant features for VLM integration~\citep{zhou2025DSR-GSM}. GeoSR improves geometric evidence use through geometry-unleashing masking and geometry-guided fusion, while employing separate models for static and dynamic reasoning~\citep{zhang2026geosr}. Building on GeoThinker’s semantic-guided intermediate-layer integration~\citep{li2026thinking}, we introduce Spatial-Dynamic Fusion (SDF) to fuse complementary spatial and dynamic priors within a single VLM, adaptively balancing their contributions.

\vspace{-1mm}
\subsection{Capability Preservation during Post-Training}
\vspace{-2mm}

Specialized post-training can improve target performance while degrading previously acquired capabilities, a phenomenon observed in language models and other adaptation settings, including embodied models~\citep{kotha2024forgetting,hancock2026actions}.
Weight averaging provides one approach to balancing adaptation and capability retention: WiSE-FT interpolates pretrained and fine-tuned checkpoints to improve robustness~\citep{wortsman2022wiseft}, while SWA~\citep{izmailov2018averaging} and EMA~\citep{moralesbrotons2024ema} aggregate parameter states within a training run to improve generalization. Prior work has examined how EMA hyperparameters should adapt to training scale, such as batch size~\citep{busbridge2023scaleema}.
In spatial post-training, however, we find that capability interference is unevenly distributed across parameter groups, calling for more targeted control over the balance between static and dynamic capabilities. Motivated by this observation, D-EMA applies configurable group-wise averaging horizons, normalized by training duration, to mitigate interference while retaining sufficient flexibility for capability enhancement across training scales.

\vspace{-1mm}
\section{Diagnosing Spatial--Dynamic Interference}
\label{sec:motivation}
\vspace{-2mm}

\begin{figure}[t]
  \centering
    \includegraphics[width=\textwidth]{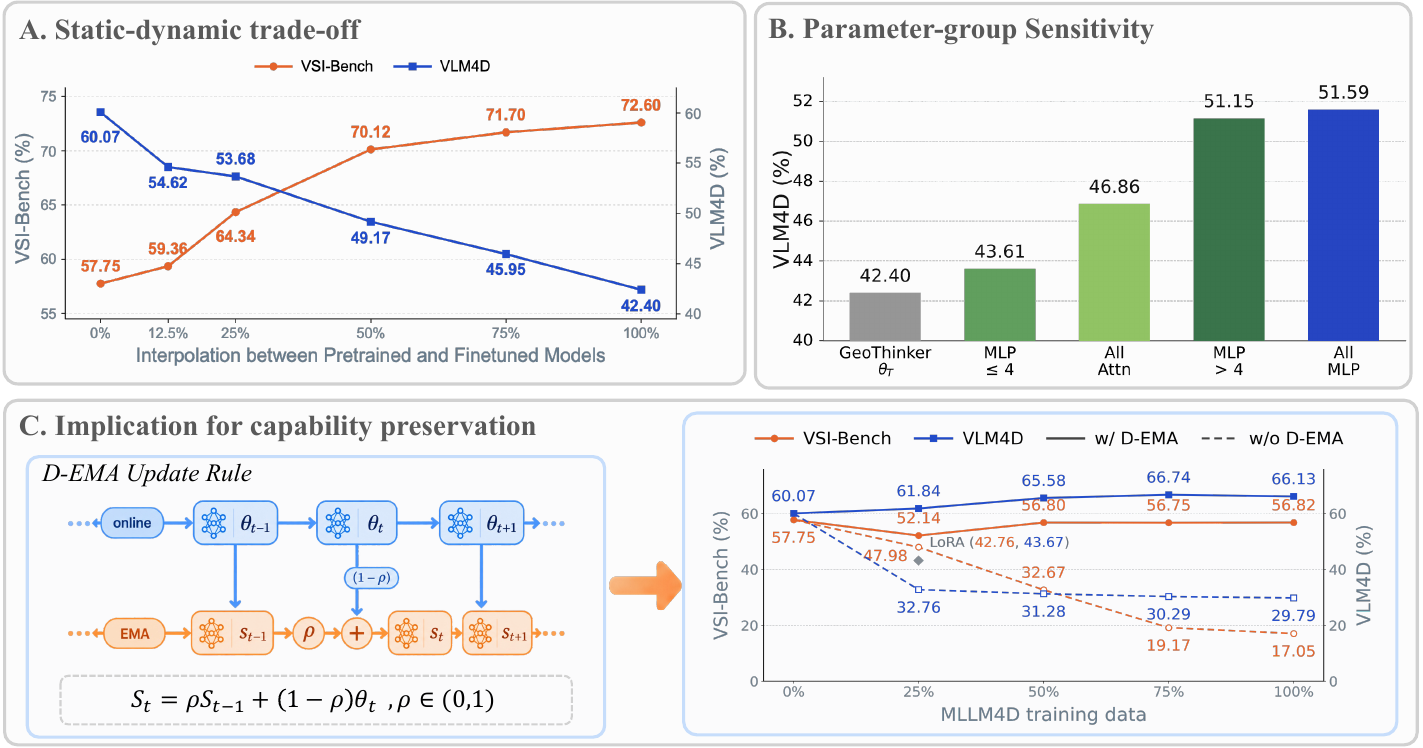} 
    \caption{
\textbf{Spatial--dynamic interference and capability preservation.}
\textbf{(A)} Interpolation between Qwen3-VL-8B and its 3D-specialized counterpart~\citep{li2026thinking} improves VSI-Bench but degrades VLM4D, revealing a clear spatial--dynamic interference.
\textbf{(B)} 
Restoring selected parameter groups toward their pretrained weights reveals that dynamic performance is particularly sensitive to post-training changes beyond the first four MLPs ($j>4$).
\textbf{(C)} D-EMA uses configurable temporal averaging to consolidate interference-sensitive pretrained parameters during adaptation. As MLLM4D~\citep{yin2026mllm4d} training data scales up, D-EMA consistently improves both static 3D and dynamic 4D performance, whereas training without D-EMA leads to degradation in both.
    }
    \label{fig:motivation}
    \vspace{-6mm}
\end{figure}

\paragraph{Static--dynamic trade-off.}To understand why existing specialization struggles to achieve unified spatial reasoning, we first examine how 3D-focused post-training reshapes the balance between static and dynamic capabilities in a pretrained VLM.
As shown in Figure~\ref{fig:motivation} (A), interpolating between the Qwen3-VL-8B~\citep{qwen3vl2025} model and its 3D-specialized counterpart GeoThinker~\citep{li2026thinking} reveals opposing trends: performance on VSI-Bench steadily improves, while performance on VLM4D consistently declines.
Notably, static 3D performance begins to saturate at intermediate interpolation coefficients, whereas dynamic 4D performance continues to decline toward the 3D-specialized checkpoint. This opposing trajectory reveals spatial--dynamic interference during post-training: improving static 3D reasoning can progressively erode dynamic competence inherited from pretraining.

\paragraph{Parameter-group sensitivity.}This observation raises a central question: which parameters drive this interference, and can dynamic competence be preserved without constraining static 3D adaptation?
To answer this, we conduct a parameter-group sensitivity analysis by selectively restoring different modules and layer ranges in the 3D-specialized model toward their pretrained values and measuring the resulting changes in static and dynamic performance.
As shown in Figure~\ref{fig:motivation} (B), the effect varies substantially across parameter groups: restoring late-layer MLPs recovers substantially more dynamic performance than other groups, but also partially reverses the static 3D gains acquired during post-training. This contrasting response indicates that spatial--dynamic interference is particularly pronounced in late-layer MLPs.
These observations motivate configurable temporal consolidation across parameter groups.
Interference-sensitive parameters should be preserved more strongly, while less sensitive parameters should retain greater flexibility for capability enhancement. Detailed analyses of parameter sensitivity are provided in Appendices~\ref{sec:mlp_layer_boundary} and~\ref{sec:weight_interpolation}.

 \paragraph{Implications for capability preservation.}These findings motivate a selective preservation strategy that can constrain interference-sensitive parameters without restricting the overall adaptation process. Temporal weight averaging provides a soft mechanism for this purpose. Unlike hard freezing, it consolidates parameter states throughout training while leaving the online adaptation process unconstrained, allowing the model to adapt while retaining influence from earlier states. 
Standard EMA, however, applies a shared averaging horizon to all parameters and therefore does not account for their different sensitivities to specialization.
We therefore introduce \textbf{Decoupled Exponential Moving Average (D-EMA)}, which assigns configurable averaging horizons to pretrained parameter groups, allowing stronger consolidation to more sensitive groups while allowing others greater flexibility in Figure~\ref{fig:motivation}(C). The formulation is detailed in Section~\ref{sec:depart_ema}. 

Figure~\ref{fig:motivation}(C) further shows that D-EMA remains effective as post-training scales.
Without D-EMA, increasing the amount of post-training data leads to progressively greater degradation in dynamic reasoning. In contrast, D-EMA preserves dynamic competence while allowing static 3D performance to continue benefiting from additional data.
This suggests that spatial--dynamic interference can intensify with training scale and that group-wise temporal averaging remains effective across data regimes. 
D-EMA thus provides optimization-side capability preservation while retaining sufficient flexibility for joint spatial--dynamic enhancement.

\vspace{-3mm}
\section{Method}
\label{sec:method}
\vspace{-2mm}

\begin{figure}[t]
  \centering
    \includegraphics[width=\textwidth]{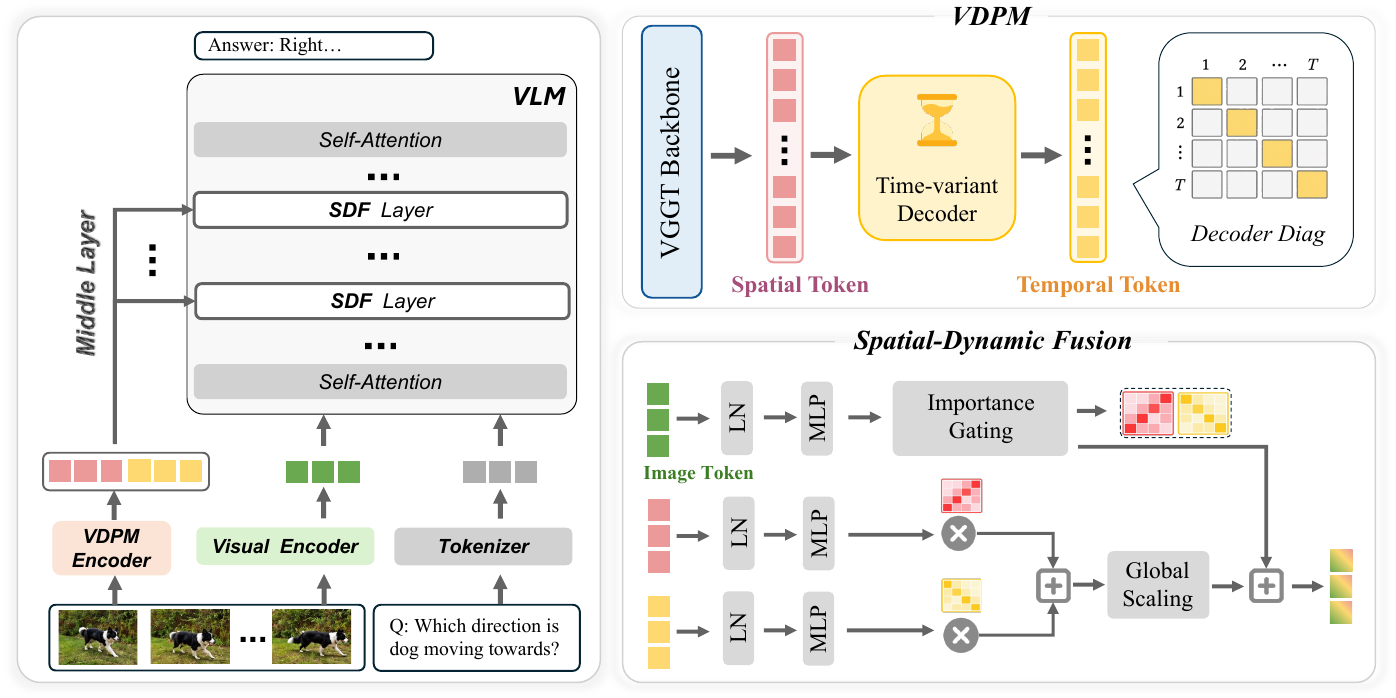} 
    \caption{\textbf{Overview of the \name architecture.} Our framework integrates complementary spatial and dynamic priors from a dynamic geometry encoder through Spatial-Dynamic Fusion (SDF) layers. Importance Gating uses the VLM's semantic context to selectively modulate both feature streams, enabling joint reasoning about geometric structure and temporal dynamics.
    }
    \label{fig:model_overview}
    \vspace{-4mm}
\end{figure}

\name unifies static 3D and dynamic 4D reasoning through two complementary components: Spatial-Dynamic Fusion (SDF) for joint capability enhancement and Decoupled Exponential Moving Average (D-EMA) for interference-aware capability preservation.
As illustrated in Figure~\ref{fig:model_overview}, SDF integrates complementary spatial and temporal features from a dynamic geometry encoder into intermediate VLM representations, adaptively modulated by the VLM's semantic context. During post-training, D-EMA applies group-specific temporal averaging to pretrained parameters, balancing capability preservation with sufficient flexibility for adaptation.
The model architecture, SDF, and D-EMA are detailed in Sections~\ref{sec:method_arch}, \ref{sec:method_fusion}, and \ref{sec:depart_ema}, respectively.

\vspace{-2mm}
\subsection{Model Architecture}
\label{sec:method_arch}
\vspace{-1mm}

\textbf{Preliminaries.}
Given a sequence of RGB images $\mathcal{I}=\{I_i\}_{i=1}^{n}$ and a natural language query $Q$, our model generates a textual response using a Qwen-VL backbone.
The visual encoder extracts patch features, which are spatially merged into visual tokens and jointly processed with the query tokens by the language model.
Following the Qwen-VL series~\citep{Qwen2.5-VL, Qwen3-VL},
each $s\times s$ block of adjacent patch features is merged into one visual token.
For frames of size $h\times w$ and vision patch size $p_s$, the resulting token grid contains
$L=\lfloor h/(sp_s)\rfloor \times \lfloor w/(sp_s)\rfloor$
tokens per frame. 
At an interaction layer $j$, we organize the image token hidden states as
$\mathbf{H}_j^{\mathrm{img}}\in\mathbb{R}^{n\times L\times c}$,
where $c$ is the LLM hidden dimension. These representations provide the semantic context for SDF.

\textbf{Dynamic Geometry Encoder.}
To provide geometric priors for both spatial structure and temporal dynamics, we adopt VDPM~\citep{sucar2026v(VDPM)} as our dynamic geometry encoder. As shown in Figure~\ref{fig:model_overview}, VDPM consists of a VGGT-based backbone~\citep{wang2025vggt} and a time-variant decoder. We extract intermediate features from both components: encoder features provide spatial geometry priors, while decoder features capture temporal variations across the sequence. Together, these streams supply complementary spatial and dynamic information for reasoning.

\textbf{Efficient Feature Extraction and Alignment.}
To reduce the sequential cost of the VDPM decoder, we densely encode all $n$ input frames but decode only a temporally ordered subset $\mathcal{K}=\{k_1,\ldots,k_m\}$, where $m\leq n$. Let $E$ and $D$ denote the VDPM encoder and decoder, respectively:
\begin{equation}
\mathbf{F}_{\mathrm{enc}}=E(\mathcal{I}),
\qquad
\mathbf{F}_{\mathrm{dec}}^{\mathrm{sparse}}
=D\!\left(\mathbf{F}_{\mathrm{enc}}[\mathcal{K}]\right).
\end{equation}
We then recover frame-aligned decoder features through temporal interpolation:
\begin{equation}
\mathbf{F}_{\mathrm{dec}}
=
\operatorname{LinearInterp}_{\mathrm{time}}
\left(
\mathbf{F}_{\mathrm{dec}}^{\mathrm{sparse}};
\mathcal{K},n
\right).
\end{equation}
This design preserves dense encoder coverage while reducing the sequential decoder length from $n$ to $m$. Both feature streams are subsequently resampled to the VLM's visual-token grid, yielding
$\widetilde{\mathbf{F}}_{\mathrm{enc}}\in\mathbb{R}^{n\times L\times d_{\mathrm{enc}}}$
and
$\widetilde{\mathbf{F}}_{\mathrm{dec}}\in\mathbb{R}^{n\times L\times d_{\mathrm{dec}}}$.
The resulting token aligned spatial and dynamic priors are projected into the LLM hidden space and integrated with $\mathbf{H}_j^{\mathrm{img}}$ by SDF at selected intermediate layers. Detailed sampling and compression configurations are provided in Appendix~\ref{sec:appendix_implementation_details}.

\begin{table*}[t]
  \caption{\textbf{Cross-benchmark comparison.} AVG is the unweighted mean across all ten metrics across nine benchmarks and is reported only for models with complete coverage. 
  Bold marks our column-best results among the listed open-source models.
  $\dagger$ indicates evaluation without token compression.
  * indicates trained with S1+S2 dataset setting from VG-LLM. Benchmarks include VSI-Bench\citep{yang2025thinking}, MMSI-Bench\citep{yang2025mmsi}, MindCube\citep{yin2025spatial(mindcube)}, ViewSpatial\citep{li2025viewspatial}, SITE\citep{wang2025site}, CV-Bench~\citep{tong2024cambrian(cambrian-1)}, VLM4D~\citep{zhou2025vlm4d}, DSR~\citep{zhou2025DSR-GSM}, DSI~\citep{zhang2025dsi} (S: sample-wise, G: group-wise).
  }
  \vskip -0.1in
  \label{tab:cross-benchmark}
  \begin{center}
    \begin{small}
      \begin{sc}
        \resizebox{\textwidth}{!}{%
        \begin{tabular}{l|c|cccccc|cccc}
          \toprule

          \multirow{2}{*}{\textbf{Models}}
          & \multirow{2}{*}{\textbf{AVG.}}
          & \multicolumn{6}{c|}{Spatial statics}
          & \multicolumn{4}{c}{Temporal dynamics}\\

          &
          & \textbf{VSI}
          & \textbf{MMSI}
          & \textbf{MindCube}
          & \textbf{ViewSpatial}
          & \textbf{SITE}
          & \textbf{CV-Bench}
          & \textbf{VLM4D}
          & \textbf{DSR} & \textbf{DSI-S} & \textbf{DSI-G}\\

          \midrule
          Human & - & 79.2 & 97.2 & 94.5 & - & 67.5 & - & 98.8 & - & - & - \\
          Random Choice & - & 34.0 & 25.0 & 33.0 & 26.3 & 0.0 & - & 24.1 & - & 25.0 & 0.1 \\
          \midrule
          \rowcolor{black!9}
          \multicolumn{12}{l}{\textbf{Proprietary Models}} \\
          Seed-1.6~\citep{bytedance2025seed16}
             & - & 49.9 & 38.3 & 48.7 & 43.8 & 54.6 & 85.2 & - & - & 41.3 & 16.1 \\
          Gemini-2.5-Pro~\citep{comanici2025gemini25}
             & 50.6 & 53.5 & 38.0 & 57.6 & 46.0 & 57.0 & 85.9 & 62.0 & 31.7 & 46.9 & 27.1 \\
          GPT-5\citep{openai_gpt5_systemcard}
             & 50.5 & 55.0 & 41.8 & 56.3 & 45.5 & 61.8 & 84.6 & 67.2 & 30.8 & 40.1& 21.9 \\
        GPT-5.6 SOL\citep{openai2026gpt56sol}
             & 59.5 & 61.2 & 47.4 & 64.3 & 49.8 & 76.7 & 85.8 & 79.3 & 29.1 & 58.8 & 42.2 \\
          \midrule
          \rowcolor{black!9}
          \multicolumn{12}{l}{\textbf{Open-source General Models}} \\
          Bagel-7B-MoT\citep{deng2025bagel}
             & 37.3 & 31.4 & 31.0 & 34.7 & 41.3 & 37.0 & 76.0 & 44.1  &26.5  & 36.9 & 13.6 \\
          Qwen2.5-VL-3B-Instruct\citep{Qwen2.5-VL}
             & 36.5 & 28.6 & 28.6 & 37.6 & 31.9 & 33.1 & 71.8 & 46.3 & 28.8 & 36.6 & 21.9 \\
          Qwen2.5-VL-7B-Instruct\citep{Qwen2.5-VL}
             & 37.7 & 29.3 & 26.8 & 36.0 & 36.8 & 37.6 & 75.4 & 50.7 & 27.1 & 38.0 & 19.1 \\
          Qwen3-VL-2B-Instruct\citep{Qwen3-VL}
             & 32.6 & 49.4 & 11.9 & 31.4 & 34.2 & 35.6 & 78.4 & 42.7 & 20.9 & 16.2 & 5.5 \\
          Qwen3-VL-8B-Instruct\citep{Qwen3-VL}
             & 44.0 & 57.7 & 28.8 & 29.8 & 39.0 & 45.8 & 85.1 & 60.0 & 28.7 & 42.3 & 22.4 \\
          InternVL3-8B \citep{zhu2025internvl3}
             & 40.2 & 42.1 & 28.0 & 41.5 & 38.6 & 41.1 & 81.0 & 55.9 & 22.8 & 37.8 & 13.1 \\
          \midrule
          \rowcolor{black!9}
          \multicolumn{12}{l}{\textbf{Open-source 3D-specialized Models}} \\
          VST-7B-SFT\citep{yang2026visual}
            & 43.2 & 60.6 & 32.0 & 39.7 & 50.5 & 39.6 & 85.5 & 52.0 & 24.1 & 36.9 & 10.8 \\
          Cambrian-S-7B \citep{yang2025cambrian}
             & 41.9 & 67.5 & 25.8 & 39.6 & 40.9 & 33.0 & 76.9 & 50.4 & 35.5 & 27.4  & 22.4 \\
          Sensenova-SI-1.3-InternVL3-8B\citep{cai2025scaling(sensenova)} & 52.2 & 68.6 & \textbf{42.5} & 89.9 & \textbf{61.3} & 47.5 & 83.0 &  50.3 & 34.6 & 36.1 & 8.3 \\
            
          VLM-3R-7B\citep{fan2025vlm(vlm3r)}
            & 40.3 & 60.9 & 27.9 & 40.0 & 40.5 & 31.3 & 71.8 & 52.0 & 30.5 & 36.9 & 10.8 \\
          VG-LLM-8B\citep{zheng2025learning(vgllm)} & 42.8 & 49.6 & 28.4 & 32.7 & 42.9 & 52.6 & 82.7 & 46.5 & 38.4 & 38.7 & 15.9 \\
          VG-LLM-8B*\citep{zheng2025learning(vgllm)} & 44.4 & 62.2 & 30.0 & 36.1 & 45.8 & 50.5 & 81.7 & 48.8 & 39.0 & 35.7 & 13.8 \\

          4DThinker\textsubscript{Qwen2.5-3B}\citep{chen20264dthinker} & 25.1 & 17.8 & 9.24 & 28.7 & 15.6  & 31.8  & 46.1 & 39.3 & 9.3 & 31.8 & 21.2 \\

            GeoThinker\textsubscript{Qwen2.5-VL-7B}~\citep{li2026thinking} & 48.9 & 68.5 & 31.7 & 83.6 & 41.9 & 54.8 & 82.1 & 45.8 & 33.5 & 35.3 & 11.5 \\
            
            GeoThinker\textsubscript{Qwen3-VL-8B}~\citep{li2026thinking} & 49.1 & \textbf{72.6} & 30.9 & 83.0 & 45.9 & 55.9 & 85.1 & 42.4 & 30.9 & 33.0 & 11.0 \\
            
            \midrule
            \rowcolor{black!9}
            \multicolumn{12}{l}{\textbf{Open-source 4D-specialized Models}} \\

            MLLM-4D-RFT\textsubscript{Qwen3-VL-8B}~\citep{yin2026mllm4d} & 45.7 & 55.5 & 27.4 & 27.5 & 42.2 & 58.2 & 85.6 & 61.0 & 37.8 & 39.4 & 22.1 \\
            GeoSR-4D~\citep{zhang2026geosr} & 42.6 & 32.9 & 23.9 & 42.5 & 37.4 & 48.7 & 76.3 & 42.7 & 66.5 & 39.0 & 15.8 \\
            DSR-GSM~\citep{zhou2025DSR-GSM} & 42.8 & 33.5 & 25.4 & 39.2 & 36.9 & 50.7 & 78.2 & 48.4 & 62.2 & 39.2 & 13.8 \\

            \midrule
            \rowcolor{navyblue!9}
            \multicolumn{12}{l}{\textbf{Ours}} \\

            \name\textsubscript{Qwen3-VL-8B} & 60.3 & 69.3 & 31.5 & 90.3 & 48.0 & 59.4 & \sota{87.1} & 60.0 & \sota{67.7} & 50.5 & 39.1 \\
            \name\textsubscript{Qwen3-VL-8B} $\dagger$ & \sota{60.5} & 67.8 & 31.5 & \sota{90.4} & 47.4 & \sota{59.6} & \sota{87.1} & \sota{62.3} & 67.2 & \sota{51.2} & \sota{40.1} \\

          \bottomrule
        \end{tabular}}
      \end{sc}
    \end{small}
  \end{center}
  \vspace{-5mm}
\end{table*}

\subsection{Spatial-Dynamic Fusion}
\label{sec:method_fusion}

To incorporate complementary spatial and dynamic priors into the VLM, we introduce Spatial-Dynamic Fusion (SDF). As illustrated in Figure~\ref{fig:model_overview}, SDF uses intermediate semantic representations to modulate the contribution of each prior stream through spatial and dynamic importance gating, together with global scaling. 

\textbf{Importance Gating.}
Spatial and dynamic evidence may contribute differently across visual regions and reasoning contexts. At an interaction layer $j$, we predict separate importance scores from the image token hidden states $\mathbf{H}_j^{\mathrm{img}}$:
\begin{equation}
S_j^{\mathrm{s}}
=
\operatorname{Sigmoid}\!\left(
\operatorname{MLP}_{j}^{\mathrm{s}}\!\left(
\operatorname{LN}(\mathbf{H}_j^{\mathrm{img}})
\right)
\right),
\qquad
S_j^{\mathrm{d}}
=
\operatorname{Sigmoid}\!\left(
\operatorname{MLP}_{j}^{\mathrm{d}}\!\left(
\operatorname{LN}(\mathbf{H}_j^{\mathrm{img}})
\right)
\right).
\end{equation}
where $S_j^{\mathrm{s}},S_j^{\mathrm{d}}\in\mathbb{R}^{n\times L\times1}$ denote spatial and dynamic importance maps, respectively. Each importance network is a lightweight bottleneck MLP that predicts a scalar weight for each visual token. The two importance maps independently modulate their corresponding prior streams, allowing semantic representations to emphasize relevant evidence at each aligned location.

\textbf{Dual-Stream Combination.}
Unlike prior approaches~\citep{li2026thinking} that rely on cross-attention for feature fusion, we directly add the contributions of the two aligned streams without cross-attention. The aligned encoder and decoder features are separately normalized with LayerNorm, modulated by their respective importance scores, and projected into the LLM hidden space:
\begin{equation}
\mathbf{F}_{j}^{\mathrm{fused}}
=
\mathbf{W}_{\mathrm{os}}
\left(
\operatorname{LN}
\left(
\widetilde{\mathbf{F}}_{\mathrm{enc}}
\right)
\odot S_j^{\mathrm{s}}
\right)
+
\mathbf{W}_{\mathrm{od}}
\left(
\operatorname{LN}
\left(
\widetilde{\mathbf{F}}_{\mathrm{dec}}
\right)
\odot S_j^{\mathrm{d}}
\right),
\end{equation}
where
$\mathbf{W}_{\mathrm{os}}\in\mathbb{R}^{c\times d_{\mathrm{enc}}}$
and
$\mathbf{W}_{\mathrm{od}}\in\mathbb{R}^{c\times d_{\mathrm{dec}}}$
are learned projections. The importance scores are broadcast along the feature dimension. Since both streams share the frame and spatial token grid of $\mathbf{H}_j^{\mathrm{img}}$, their contributions are combined at corresponding locations, yielding
$\mathbf{F}_{j}^{\mathrm{fused}}\in\mathbb{R}^{n\times L\times c}$.

\textbf{Residual Scaling and Deployment.}
We inject the combined prior into the visual hidden states through a learnable scalar gate $g$ following \citep{li2026thinking}:
\begin{equation}
\mathbf{H}_{j,\mathrm{out}}^{\mathrm{img}}
=
\mathbf{H}_j^{\mathrm{img}}
+
\tanh(g)\,
\mathbf{F}_{j}^{\mathrm{fused}}.
\end{equation}
The gate is initialized to zero, making SDF an identity mapping at initialization. During training, the model jointly learns the global injection strength and token-level importance maps. The importance gating controls where each prior contributes, while residual scaling controls the overall magnitude of the contribution.
We deploy SDF at selected intermediate LLM layers. At each interaction layer, SDF updates the image token hidden states, which continue through the backbone alongside the text token states for subsequent multimodal reasoning. Since the priors are injected directly into existing visual representations, SDF introduces no additional tokens to the LLM sequence.

\begin{table*}[t]
  \caption{
  \textbf{Performance comparison on ReVSI-Bench.}
ReVSI adapts ground-truth answers to the evaluated frame count, assessing spatial reasoning under different frame budgets. \name achieves the highest overall accuracy among the compared open-source models.
  }
  \vspace{-2mm}
  \label{tab:revsi_comparisons}
  \begin{center}
    \begin{small}
      \begin{sc}
      \setlength{\tabcolsep}{3pt}
        \resizebox{\textwidth}{!}{%
        \begin{tabular}{lc|c|cccc|ccc}
          \toprule
          \multirow{2}{*}{Methods}& \multirow{2}{*}{Frames} & \multirow{2}{*}{Avg.}
          & \multicolumn{4}{c|}{\cellcolor{orange!10}Numerical Answer}
          & \multicolumn{3}{c}{\cellcolor{yellow!10}Multiple-Choice Answer} \\
          \cmidrule(lr){4-7}\cmidrule(lr){8-10}
          & &&  Obj.Count & Abs. Dist. & Obj. Size & Room Size & Rel. Dist. & Rel. Dir. & Route Plan  \\
          \midrule
          \rowcolor{black!11}
          \multicolumn{10}{l}{\textbf{Proprietary Models}} \\
          GPT-5.2~\citep{openai2025gpt52} & 64 & 50.9 & 56.2 & 41.5 & 73.9 & \textbf{63.0} & 48.4 & 34.9 & 38.2 \\
          Gemini 3 Flash~\citep{gemini2025flash3} & 1 FPS & 57.6 & \textbf{65.7} & 53.1 & 77.6 & 52.8 & 64.6 & 47.9 & 41.8 \\
          Gemini 3 Pro~\citep{gemini_3_pro_systemcard} & 1 FPS & \textbf{60.9} & 60.1 & 54.7 & \textbf{79.3} & 51.9 & \textbf{68.1} & \textbf{56.0} & \textbf{56.4} \\

          \midrule
          \rowcolor{black!8}
          \multicolumn{10}{l}
          {\textbf{Open-source Models}} \\
          Qwen3-VL-8B-Instruct~\citep{Qwen3-VL} & 64 & 49.1 & 40.4 & 52.3 & 69.0 & 45.1 & 57.1 & 39.5 & 40.5 \\
          Qwen2.5-VL-7B-Instruct~\citep{Qwen2.5-VL} & 4 FPS & 32.6 & 36.9 & 15.0 & 49.7 & 29.0 & 31.5 & 29.5 & 36.7 \\
          InternVL3.5-8B~\citep{wang2025internvl3.5} & 64 & 47.9 & 43.3 & 54.6 & 64.2 & 47.6 & 45.0 & 36.3 & 44.4 \\
          LLaVA-Video-7B-Qwen2~\citep{zhang2024video} & 64 & 30.3 & 31.3 & 1.4 & 52.5 & 16.7 & 38.3 & 33.3 & 38.4 \\
          Cambrian-\emph{S}\xspace-7B~\citep{yang2025cambrian} & 128 & 49.1 & \textbf{48.4} & 60.5 & 65.5 & 46.7 & 37.1 & 48.5 & 37.0 \\
          VST-7B-SFT~\citep{yang2026visual} & 4 FPS & 46.4 & 35.4 & 52.6 & 67.9 & 47.2 & 49.2 & 36.9 & 35.4 \\
          Cambrian-\emph{P}\xspace\ (w/o Pose)~\citep{yang2026cambrian} & 128 & 50.1 & 42.3 & 64.6 & 64.5 & 47.4 & 38.1 & 48.3 & 45.3 \\
          Cambrian-\emph{P}\xspace~\citep{yang2026cambrian} & 128 & 52.0 & 41.4 & 68.3 & 66.7 & 45.9 & 40.2 & 48.7 & \textbf{53.1} \\
          GeoThinker\textsubscript{Qwen3-VL-8B}~\citep{li2026thinking} & 128 & 55.9 & 37.9 & \sota{74.7} & 67.6 & 57.1 & 61.8 & 47.8 & 44.7 \\

        \midrule
          \rowcolor{navyblue!11}
          \multicolumn{10}{l}{\textbf{Ours}} \\

          \name\textsubscript{Qwen3-VL-8B} & 128 & \sota{56.0} & 37.6 & 67.0 & \sota{70.8} & \sota{59.2} & \sota{64.7} & \sota{48.9} & 44.0 \\

          \bottomrule
        \end{tabular}%
        }
      \end{sc}
    \end{small}
  \end{center}
  \vskip -0.2in
\end{table*}

\subsection{Decoupled Exponential Moving Average}
\label{sec:depart_ema}

D-EMA consolidates the post-training trajectory through group-wise temporal averaging without constraining online optimization. It decouples the consolidation of pretrained parameters from fusion module adaptation by averaging the former while retaining learned weights of the latter. Motivated by the parameter-group sensitivity identified in Section~\ref{sec:motivation}, it assigns configurable averaging horizons to pretrained parameter groups according to their sensitivity to spatial specialization.

\textbf{Parameter Groups.}
We use a boundary $b$ to divide the LLM's MLP blocks into early ($j\leq b$) and late ($j>b$) groups. The sensitive group $\mathcal{G}_{\mathrm{sens}}(b)$ contains trainable pretrained MLP parameters in the late group. The base group $\mathcal{G}_{\mathrm{base}}$ contains all other trainable pretrained parameters, including early MLP blocks, attention modules, and trainable visual components. Finally, $\mathcal{G}_{\mathrm{new}}$ contains all newly introduced trainable parameters, including SDF. %

\textbf{Group-Wise Averaging.}
Let $\theta_t^{(k)}$ denote the online parameters of group $k$ observed at the $t$-th EMA update. The online model is optimized normally, while D-EMA maintains the corresponding shadow parameters as
\begin{equation}
\bar{\theta}_t^{(k)}
=
\rho_k\bar{\theta}_{t-1}^{(k)}
+
(1-\rho_k)\theta_t^{(k)},
\qquad
\bar{\theta}_0^{(k)}=\theta_0^{(k)}.
\end{equation}
For pretrained groups, $\theta_0^{(k)}$ is initialized from the pretrained checkpoint. We set
\begin{equation}
0<\rho_{\mathrm{base}}\le\rho_{\mathrm{sens}}<1,
\qquad
\rho_{\mathrm{new}}=0.
\end{equation}
A larger decay retains greater influence from earlier parameter states, enabling stronger consolidation for the sensitive group when $\rho_{\mathrm{sens}}>\rho_{\mathrm{base}}$.
Newly introduced parameters are excluded from temporal averaging because $\rho_{\mathrm{new}}=0$ gives $\bar{\theta}_t^{(\mathrm{new})}=\theta_t^{(\mathrm{new})}$. After training, the consolidated parameters $\bar{\theta}_T$ are used for evaluation and inference without changing the model architecture.

\textbf{Scale-Normalized D-EMA Horizons.}
To maintain a consistent averaging horizon relative to training duration,
we parameterize the decay for each pretrained parameter group $k$ as
\begin{equation}
\lambda_k=-N_{\mathrm{ref}}\log\rho_k,
\qquad
\rho_k'=\exp\!\left(-\frac{\lambda_k}{N_{\mathrm{target}}}\right),
\end{equation}
where $N_{\mathrm{ref}}$ and $N_{\mathrm{target}}$ denote the numbers of
D-EMA updates in the reference and target training schedules, respectively.
This conversion preserves the contribution of the initial weights at the
end of training:
$(\rho_k')^{N_{\mathrm{target}}}=\rho_k^{N_{\mathrm{ref}}}=e^{-\lambda_k}.$
Newly introduced parameters are excluded from D-EMA.

\vspace{-1mm}
\section{Experiment}
\label{sec:experiments}
\vspace{-1mm}

In this section, we first introduce the implementation details and training datasets, followed by evaluations on spatial and dynamic reasoning benchmarks in Section~\ref{exp:spatial_reasoning}. We then present ablation studies and scaling behavior in Section~\ref{exp:ablation}.
Additional controlled experiments and downstream evaluations are presented in the Appendix \ref{sec:appendix_spatial_reasoning}, \ref{sec:appendix_embodied_manipulation} and \ref{sec:appendix_temporal_dynamics}.

\textbf{Implementation Details.}
We train \name with a batch size of 64 and a learning rate of $1\times10^{-5}$, using 32 input frames per scene and $m=8$ decoded frames in VDPM. The fusion ratio is set to $r_{\mathrm{fusion}}=0.5$, and the D-EMA layer boundary is $b=4$. To accommodate broader adaptation during scaled training, we use a D-EMA decay of $0.999834$ for pretrained parameters, while newly introduced modules are excluded from averaging. D-EMA starts after 150 optimization steps and updates every 10 steps. Training uses 16 PPU-ZW810E accelerators with 96GB memory each. 

\textbf{Training Data.}
Our training mixture contains approximately 2.68M samples, extending GeoThinker's scaled data recipe~\citep{li2026thinking}. It includes sampled SPAR-7M~\citep{zhang2025flatland(spar)} and the LLaVA-Hound split of LLaVA-Video-178K \citep{zhang2024video}, spatial instruction data from VLM-3R \citep{fan2025vlm(vlm3r)}, VSI-590K \citep{yang2025cambrian}, and MindCube \citep{yin2025spatial(mindcube)}, and 430k general video samples from~\citep{yang2025cambrian}. The mixture also includes 13K DSR~\citep{zhou2025DSR-GSM} samples associated with 2,000 videos, covering 26\% of its official training data, and 1M MLLM4D samples from ~\citep{yin2026mllm4d}, covering 50\% of its training set. %

\subsection{Evaluation Results}
\label{exp:spatial_reasoning}
We conduct evaluation across multiple benchmarks, including (1) Spatial statics: ReVSI~\citep{zhang2026revsi}, VSI-Bench~\citep{yang2025thinking}, MMSI-Bench~\citep{yang2025mmsi}, MindCube-Tiny~\citep{yin2025spatial(mindcube)}, ViewSpatial~\citep{li2025viewspatial}, SITE~\citep{wang2025site}, CV-Bench~\citep{tong2024cambrian(cambrian-1)} and (2) Temporal dynamics: VLM4D~\citep{zhou2025vlm4d}, DSR-Bench~\citep{zhou2025DSR-GSM}, DSI-Bench~\citep{zhang2025dsi} with sample-wise and group-wise protocol.

\textbf{Cross-benchmark evaluation.}
Table~\ref{tab:cross-benchmark} evaluates static 3D and dynamic 4D reasoning across a broad benchmark suite. 
Our 8B model achieves the highest average score of 60.3, outperforming all compared open-source models.
Compared with 3D-specialized models, \name substantially strengthens dynamic 4D reasoning while retaining strong static 3D performance. It outperforms GeoThinker, which shares the same Qwen3-VL-8B backbone, by 11.2 points on average and improves all four dynamic metrics. In particular, VLM4D accuracy rises from 42.4 to 60.0, recovering the pretrained backbone's performance, while DSR-Bench increases from 30.9 to 67.7.
Compared with 4D-specialized models, including MLLM-4D-RFT, GeoSR-4D, and DSR-GSM, \name achieves higher scores across static and dynamic metrics, showing that dynamic gains need not sacrifice static 3D capability. Relative to general VLMs, \name raises the pretrained Qwen3-VL-8B average from 44.0 to 60.3 and surpasses GPT-5.6 Sol by 0.8 points. Overall, \name achieves a stronger and more balanced static--dynamic capability within a unified model.

\textbf{Spatial reasoning on ReVSI.}
To complement the cross-benchmark comparison, we evaluate \name on ReVSI in Table~\ref{tab:revsi_comparisons}, which adapts ground-truth answers to the evaluated frame count.
Under the same 128-frame budget, \name surpasses Cambrian-P by 4.0 percentage points and slightly improves upon GeoThinker's 55.9\%, with gains in object and room size estimation, relative distance, and relative direction.
These results demonstrate that \name maintains strong 3D spatial reasoning alongside its improved dynamic capabilities.

\textbf{Effect of inference-time token compression.}
Disabling token compression at inference yields a modest increase in the average score, from 60.3 to 60.5. VLM4D improves from 60.0 to 62.3, and scores under both DSI evaluation protocols also increase, whereas VSI-Bench decreases from 69.3 to 67.8. These results suggest that retaining more visual tokens can benefit dynamic reasoning, but the gains are task-dependent and do not consistently translate to better overall performance.

\subsection{Ablation Study}
\label{exp:ablation}

\textbf{Component Ablations and Model Scaling.}
Table~\ref{tab:ablation}(a) incrementally builds DynThinker upon GeoThinker under the same data recipe and training schedule with the Qwen3-VL-4B backbone.
Adding D-EMA to GeoThinker improves VSI-Bench from 51.31 to 53.14 and VLM4D from 45.20 to 53.57, recovering much of the dynamic performance lost during specialization while also improving static spatial reasoning.
This substantial dynamic recovery shows that D-EMA effectively mitigates the interference introduced during 3D-specialized post-training. 
Replacing the geometry encoder with VDPM provides additional gains, while incorporating encoder-side spatial priors and decoder-side temporal priors further strengthens both capabilities, with the full 4B model reaching 56.56 on VSI-Bench and 59.74 on VLM4D.
Scaling the complete model to an 8B backbone further improves these scores to 58.16 and 61.78. Overall, the ablation results highlight the complementary roles of D-EMA for capability preservation and SDF for enhancing static and dynamic reasoning.

\begin{table}[t]
\centering
\vspace{-2mm}
\caption{\textbf{Component ablations and dynamic-data scaling.}
(a) From GeoThinker with the same Qwen3-VL-4B backbone, we progressively add D-EMA, VDPM, spatial and temporal priors, then scale the backbone to 8B.
(b) MLLM4D data scaling on the VLM4D Real split, with reported MLLM-4D SFT results included for comparison. The dashed line marks the previous SOTA MLLM-4D RFT score of 63.1, while the lower panel separately reports overall VLM4D accuracy.}
\label{tab:ablation}
\vspace{-1mm}

\noindent\begin{minipage}[t]{0.6\textwidth}
\vspace{0pt}
\centering
{\fontsize{8.5}{10}\selectfont\textbf{(a) Component and model scaling}\par}
\vspace{0.4mm}
\renewcommand{\arraystretch}{1.20}\resizebox{\linewidth}{!}{%
\begin{tabular}{@{}lccccc cc@{}}
\toprule
\textbf{Model}
& \multicolumn{5}{c}{\textbf{Components}}
& \multicolumn{2}{c}{\textbf{Results}} \\
\cmidrule(lr){2-6} \cmidrule(lr){7-8}
& \textbf{SFT} & \textbf{D-EMA} & \textbf{VDPM}
& \textbf{Static} & \textbf{Tem.} & \textbf{VSI} & \textbf{VLM4D} \\

\midrule
Qwen3-VL-4B
& \xmarkg & \xmarkg & \xmarkg & \xmarkg & \xmarkg
& 54.90 & 57.10 \\
GeoThinker\textsubscript{Qwen3-VL-4B}
& \cmark & \xmarkg & \xmarkg & \xmarkg & \xmarkg
& 51.31 & 45.20 \\
\midrule
\multirow{4}{*}{\name\textsubscript{Qwen3-VL-4B}}
& \cmark & \cmark & \xmarkg & \xmarkg & \xmarkg
& 53.14 & 53.57 \\
& \cmark & \cmark & \cmark & \xmarkg & \xmarkg
& 53.60 & 54.01 \\
& \cmark & \cmark & \cmark & \cmark & \xmarkg
& 56.19 & 58.64 \\
& \cmark & \cmark & \cmark & \cmark & \cmark
& \textbf{56.56} & \textbf{59.74} \\
\midrule
\name\textsubscript{Qwen3-VL-8B}
& \cmark & \cmark & \cmark & \cmark & \cmark
& \sota{58.16} & \sota{61.78} \\
\bottomrule
\end{tabular}%
}
\end{minipage}%
\hfill%
\begin{minipage}[t]{0.37\textwidth}
\vspace{0pt}
\centering
{\fontsize{8.5}{10}\selectfont\textbf{(b) Dynamic data scaling}\par}
\vspace{0.4mm}

\includegraphics[width=\linewidth]{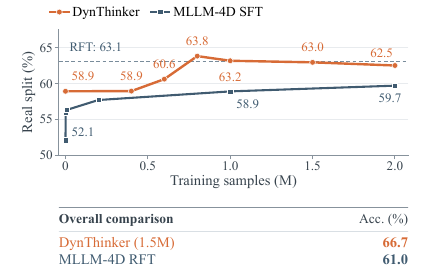}
\end{minipage}

\vspace{0pt}
\end{table}

\textbf{Dynamic-Data Scaling.}
Table~\ref{tab:ablation}(b) compares dynamic-data scaling under a controlled setting where \name and MLLM4D~\citep{yin2026mllm4d} use the same Qwen3-VL-8B backbone and training data.
On the VLM4D Real split, \name reaches 63.82 with 800K samples, surpassing MLLM-4D's reported 59.7 with 2M SFT samples and 63.1 with RFT. Performance then remains stable between 62.51 and 63.17 from 1M to 2M samples. On the full VLM4D benchmark, our 1.5M-sample model achieves 66.74, compared with 61.0 for MLLM-4D RFT. These results show that \name remains effective under dedicated 4D specialization, achieving strong dynamic reasoning with fewer SFT samples and outperforming MLLM-4D RFT on the full VLM4D benchmark.

\section{Conclusion}
\label{sec:conclusion}
We presented \name, a unified framework for static 3D and dynamic 4D spatial reasoning.
Our analysis reveals spatial--dynamic interference during post-training, with the balance between the two capabilities particularly sensitive to changes in late-layer MLPs.
To mitigate this interference, D-EMA preserves pretrained capabilities through group-wise temporal averaging, while SDF enhances spatial--dynamic reasoning by integrating complementary spatial and temporal priors into VLM representations under semantic guidance.
Across diverse benchmarks, \name achieves strong and balanced 3D and 4D performance. Ablation and scaling studies further validate the effectiveness of parameter consolidation and dual-stream feature integration, while showing that \name remains effective under dedicated 4D specialization. 
Overall, our results demonstrate that strong static 3D and dynamic 4D capabilities can be unified, advancing toward general spatial intelligence.

\newpage
\clearpage

\subsection*{AI Use Statement}

In this work, we used generative AI tools to support grammar and language editing, and the refinement of titles, narrative structure, and overall readability. We also used these tools to provide feedback on experimental design, assist with the implementation and debugging of research code, interpret experimental results, brainstorm candidate architectural variations, and suggest high-level concepts and example layouts for scientific figures. The core research hypothesis, the proposed D-EMA and SDF components, the final model and training design, and all scientific findings were formulated and determined by the authors. AI tools also assisted with literature searches and bibliographic verification. All references were reviewed by the authors. The tools were not used to generate or clean datasets, annotate or filter data, create synthetic training data, or independently formulate mathematical claims. All scientific figures were manually created and edited by the authors. Generative AI assistance was limited to creative suggestions and preliminary examples. All AI-assisted code was manually reviewed and tested through experiments, and all technical descriptions, experimental interpretations, and numerical claims were checked against the implementation and experimental results. We take responsibility for the final content of this work, including its text, claims, code, and artifacts.

\subsection*{Ethics Statement}

This work uses publicly available research datasets and pretrained models and does not involve new data collection, private data, interaction with human subjects, or additional human annotation. We use these resources in accordance with their respective licenses and terms of use. The proposed method is intended as a research prototype for static and dynamic spatial reasoning and should not be directly deployed in embodied, robotic, autonomous-driving, or other safety-critical environments without additional validation and appropriate safeguards. As with other multimodal models, its behavior may inherit biases and failure modes from the underlying pretrained models and public datasets, including inaccurate spatial judgments and incorrect interpretations of physical dynamics. We are not aware of other specific ethical concerns arising from this work.

\subsection*{Reproducibility Statement}

The main paper describes the model architecture, SDF, D-EMA, the unified post-training procedure, and the primary evaluation settings. The appendix provides additional implementation details, including model configurations, training data composition and sampling ratios, optimization objectives and hyperparameters, D-EMA parameter groups and decay settings, EMA initialization and update schedules, dynamic-frame sampling and temporal interpolation, token compression configurations, and benchmark-specific evaluation protocols. All experiments use publicly available datasets and a fixed random seed. The core model is trained in FP16 precision with DeepSpeed ZeRO Stage 2 on 16 PPU-ZW810E accelerators with 96GB of memory each, requiring approximately eight days of training. The 4D data-scaling experiments use the same backbone and training pipeline while varying the amount of 4D training data.

For VLM4D, we apply a fixed deterministic answer parser uniformly to all internally evaluated checkpoints. Its extraction rules and implementation are documented with the evaluation protocol. Baseline results are taken from the corresponding papers or public leaderboards when available, while models without published results, particularly on temporal-dynamic benchmarks, are evaluated using our implementation. The source of each reported result is identified in the paper or appendix. Upon acceptance, we will release the source code, trained model weights, training configurations, data processing scripts, deterministic answer parser, and evaluation scripts to facilitate reproduction of the reported results.

The code and models will be open-sourced upon acceptance.

\bibliographystyle{iclr2027_conference}
\bibliography{iclr2027_conference}

\clearpage

\appendix
\appendix
\clearpage
\newpage

\section{Appendix/supplemental material}
The outline of the Appendix is as follows:
\begin{itemize}
    \item Implementation details;
    \item More results under controlled data regime;
    \begin{itemize}
    \item More results under main scaled regime;
    \begin{itemize}
        \item More analysis on frame ablation;
        \item More analysis on scaling ablation;
        \item More evaluation on STI-Bench;
        \item More analysis on comparison to GeoThinker;
        \item More visualization on VLM4D;
    \end{itemize}
    \item More results under 4D scaled regime;
    \begin{itemize}
        \item More analysis on VLM4D comparison;
    \end{itemize}
    \item More results under 3D vanilla regime;
    \begin{itemize}
        \item More analysis on spatial reasoning;
        \item More ablation study on D-EMA;
    \end{itemize}
    \item More results under 4D vanilla regime;
    \begin{itemize}
        \item More analysis on temporal dynamics;
    \end{itemize}
    \end{itemize}
    \item More analysis of sensitivity analysis;
    \begin{itemize}
        \item More analysis on mlp layer boundary;
        \item More analysis on weight interpolation;
    \end{itemize}
    \item More results under downstream application;
    \begin{itemize}
        \item More analysis on embodied manipulation;
    \end{itemize}

\end{itemize}

\section{Implementation Details}
\label{sec:appendix_implementation_details}

\subsection{Model Configurations}
\textbf{Frame Sampling and Dynamic Decoding.}
Following GeoThinker~\citep{li2026thinking}, we train with 32 input frames per sequence for \emph{scaled regime} in the main paper. The VDPM encoder processes all 32 frames, while the sequential decoder processes features from 8 uniformly sampled frames during both training and inference. The resulting decoder features are linear interpolated along the temporal axis to construct representations for all 32 input frames. Both feature streams are subsequently resampled to the VLM's visual token grid before fusion.

\textbf{Visual Token Compression.}
Following GeoThinker~\citep{li2026thinking}, we employ a larger spatial merge size, increasing it from 2 to 4 for inputs containing more than 8 frames. For inputs with at most 8 frames, we retain the backbone's native merge size of 2. Thus, the input-dependent merge size is
\begin{equation}
s(n)=
\begin{cases}
2, & n\leq 8,\\
4, & n>8.
\end{cases}
\end{equation}
This configuration reduces the visual token sequence length for longer inputs while retaining a finer spatial grid for shorter inputs. The compression threshold depends on the original input frame count $n$. Consequently, our 32-frame training configuration uses spatial merge size $s=4$, with 8 frames selected for decoder processing.

\subsection{Data Curation}
\label{sec:appendix_details_data_curation}

For the scaled regime training of our model, we curated a large-scale multimodal dataset totaling 2.68M samples. The data composition is as follows:

\textbf{Spatial Reasoning:} Cambrian-S VSI-Bench instruction (590k), SPAR (234k), VLM-3R VSI-Bench instruction (205k), VLM-3R VSTI-bench instruction (132k), MindCube training set (10k).

\textbf{Temporal Dynamics}: 26\% DSR training set (13k) and 50\% MLLM4D training set (1M).

\textbf{General Video:} LLaVA-Hound (64k) and a subset of general video data sampled from Cambrian-S-3M (430k).

\section{Overview of Additional Results}

We use several experimental settings to examine complementary aspects of \name. This section summarizes their training data, model configurations, and locations in the paper.

\textbf{Main scaled regime.} The main \name model uses Qwen3-VL-8B and is trained with 32 input frames on a mixture of approximately 2.68M spatial reasoning, temporal-dynamics, and general video samples, as detailed in Section \ref{sec:experiments} and Appendix \ref{sec:appendix_details_data_curation}. Its cross-benchmark results are reported in Tables \ref{tab:cross-benchmark} and \ref{tab:revsi_comparisons}. Additional analyses of frame scaling and the scaled training configuration appear in Appendices \ref{sec:appendix_frame_ablation} and \ref{sec:appendix_scaling_ablation}. We further evaluate this model on STI-Bench in Appendix \ref{sec:appendix_sti_bench}. Appendix \ref{sec:appendix_geothinker_comparison} provides a broader comparison with GeoThinker, while Appendix \ref{sec:appendix_visualization} presents qualitative comparisons of spatial and dynamic reasoning.

\textbf{Dedicated 4D data-scaling setting}. To study adaptation to dynamic reasoning, we vary the amount of MLLM4D training data using the Qwen3-VL-8B backbone. At matched data budgets, \name and MLLM4D use the same backbone and training samples. This setting is separate from the 2.68M-sample mixture used for the main model. Results on the VLM4D Real split are shown in Table ~\ref{tab:ablation}(b), while overall VLM4D results and the evaluation protocol comparison are provided in Appendix \ref{sec:appendix_vlm4d_comparison}.

\textbf{Static 3D vanilla regime.} For controlled spatial reasoning experiments, we train with 8 input frames using 234K samples from SPAR-7M and 64K samples from LLaVA-Hound. Table 3(a) studies the incremental components starting from GeoThinker. Appendix \ref{sec:appendix_spatial_reasoning} reports the VSI-Bench comparison, while Appendix \ref{sec:appendix_edma_ablation} examines D-EMA configurations and ablations starting from direct spatial supervised fine-tuning. These two ablations therefore have different starting models, even when they report results for the same final \name configuration.

\textbf{Dynamic 4D vanilla regime.} For controlled preservation experiments, we add 200K MLLM4D samples to the 298K-sample static vanilla mixture and train with 8 input frames. Appendix \ref{sec:appendix_temporal_dynamics} compares preservation strategies on VSI-Bench and VLM4D and studies how inference-time frame count and token compression affect performance.

\textbf{Embodied setting.} To examine transfer to manipulation, we incorporate SDF into the Qwen3-VL-4B backbone of StarVLA-OFT and train a single policy across the four LIBERO suites for 30,000 steps. The setup and results are reported in Appendix \ref{sec:appendix_embodied_manipulation}.

\section{Additional results (scaled regime)}
\subsection{Further Frame ablation}
\label{sec:appendix_frame_ablation}
\begin{table}[h]
\centering
\small
\setlength{\tabcolsep}{1.5mm}
\caption{
\textbf{Comparisons of zero-shot frame scaling.} Both models use Qwen3-VL-8B and are trained with 32-frame inputs. We vary the evaluation frame budget from 16 to 128 on VSI-Bench, VSI-Debiased, and ReVSI-Bench. ReVSI-Bench adjusts ground-truth answers according to the evaluated frame count.
}
\resizebox{0.7\linewidth}{!}{
\begin{tabular}{llcccc}
\toprule
\multirow{2}{*}{\textbf{Model}} & \multirow{2}{*}{\textbf{Benchmark}} & \multicolumn{4}{c}{\textbf{\# Frames}} \\
\cmidrule(lr){3-6}
 &  & \textbf{16} & \textbf{32} & \textbf{64} & \textbf{128} \\

\midrule
\multirow{3}{*}{GeoThinker\textsubscript{Qwen3-VL-8B-32frame}}
& VSI & 69.2 & 72.6 & 73.4 & \sota{73.4} \\
& VSI-Debiased & 64.3 & 66.3 & 67.7 & \sota{68.1} \\
& ReVSI & 52.6 & 53.9 & 54.8 & \sota{55.9}\\

\midrule
\multirow{3}{*}{\name\textsubscript{Qwen3-VL-8B-32frame}}
& VSI & 64.0 & 69.3 & 70.7 & \sota{72.1} \\
& VSI-Debiased & 57.1 & 63.1 & 64.4 & \sota{66.5} \\
& ReVSI & 50.5 & 53.7 & 55.7 & \sota{56.0} \\
 
\bottomrule
\end{tabular}}
\label{tab:appendix_revis_frame}
\end{table}

Table~\ref{tab:appendix_revis_frame} evaluates generalization to frame counts beyond the 32-frame training configuration. \name improves consistently with additional evaluation frames, increasing from 50.5 to 56.0 on ReVSI-Bench between 16 and 128 frames. It surpasses GeoThinker on ReVSI-Bench from 64 frames onward. GeoThinker remains stronger on VSI-Bench and VSI-Debiased, indicating that \name's advantage is concentrated in the frame-adaptive spatial reasoning measured by ReVSI-Bench rather than uniformly across all spatial metrics.

\vspace{-3mm}
\subsection{Scaling Ablation study}
\vspace{-1mm}
\label{sec:appendix_scaling_ablation}

\begin{table*}[t]
\centering
\caption{
\textbf{Ablations of the scaled training configuration.}
All rows use \name with Qwen3-VL-8B.
The base mixture follows GeoThinker's data recipe without PhysGame~\citep{cao2024physgame} or the additional 430K general video samples, and includes 10\% of MLLM4D.
The final two rows cumulatively expand this mixture.
Target drift budget denotes the target drift budget for pretrained parameters.
}
\label{tab:appendix_scaled_ablation}
\vspace{-1mm}
\small
\setlength{\tabcolsep}{4pt}
\renewcommand{\arraystretch}{1.08}
\resizebox{\textwidth}{!}{%
\begin{tabular}{l|c|c|cccccc|cccc}
\toprule
\multirow{2}{*}{\textbf{Training Configuration}}
& \multirow{2}{*}{\shortstack{\textbf{Target D-EMA}\\\textbf{Drift}}}
& \multirow{2}{*}{\textbf{Avg.}}
& \multicolumn{6}{c|}{\textbf{Spatial Statics}}
& \multicolumn{4}{c}{\textbf{Temporal Dynamics}} \\
\cmidrule(lr){4-9}\cmidrule(lr){10-13}
& &
& \textbf{VSI}
& \textbf{MMSI}
& \textbf{MindCube}
& \textbf{ViewSpatial}
& \textbf{SITE}
& \textbf{CV-Bench}
& \textbf{VLM4D}
& \textbf{DSR}
& \textbf{DSI-S}
& \textbf{DSI-G} \\
\midrule

\multirow{2}{*}{Base mixture, 8-frame training}
& 12.5\% & 50.28
& 61.5 & 31.9 & 42.7 & 42.9 & 57.1 & 86.6
& 62.9 & 46.6 & 45.2 & 25.4 \\
& 25\% & 54.77
& 64.8 & 30.7 & 67.1 & 46.7 & 56.8 & 87.4
& 61.8 & 61.2 & 45.2 & 26.0 \\

\cmidrule(lr){1-13}
\multirow{2}{*}{Base mixture, 32-frame training}
& 25\% & 53.73
& 66.2 & 30.4 & 61.2 & 45.6 & 58.0 & 87.0
& 61.3 & 59.5 & 44.3 & 23.9 \\
& 50\% & 57.24
& 69.2 & 29.1 & 87.3 & 47.4 & 56.7 & 87.4
& 57.0 & 66.1 & 45.2 & 27.0 \\

\midrule
+ MLLM4D: $10\% \rightarrow 50\%$
& 50\% & 59.87
& 70.1 & 31.4 & 90.2 & 48.0 & 57.0 & 86.3
& 58.6 & 67.7 & 50.5 & 39.0 \\
+ 430K general video samples
& 50\% & \textbf{60.29}
& 69.3 & 31.5 & 90.3 & 48.0 & 59.4 & 87.1
& 60.0 & 67.7 & 50.5 & 39.1 \\

\bottomrule
\end{tabular}%
}
\end{table*}

Table~\ref{tab:appendix_scaled_ablation} examines the effects of the D-EMA drift budget, training frame count, and data scaling. Increasing the drift budget consistently improves the overall score by allowing greater adaptation, but introduces a trade-off across capabilities. 
With 8-frame training, increasing the target drift budget from 12.5\% to 25\% raises the average score from 50.28 to 54.77.
Under 32-frame training, increasing the budget from 25\% to 50\% raises the average from 53.73 to 57.24, driven largely by improvements on MindCube and DSR, while reducing VLM4D from 61.3 to 57.0. Expanding the MLLM4D subset from 10\% to 50\% subsequently improves the average to 59.87, with particularly strong gains on DSI-G and DSI-S. Finally, adding 430K general video samples further improves SITE from 57.0 to 59.4 and VLM4D from 58.6 to 60.0, producing the final average of 60.29. Together, these results motivate the larger drift budget and progressively expanded training mixture adopted in the final model.

\vspace{-3mm}
\subsection{Evaluation on STI-Bench}
\label{sec:appendix_sti_bench}
\vspace{-1mm}

\definecolor{iccvblue}{rgb}{0.21,0.49,0.74}
\definecolor{lightblue}{RGB}{173,216,230}
\definecolor{orange}{HTML}{FFA64D}
\definecolor{lightorange}{HTML}{FFD4A3}
\definecolor{mygreen}{RGB}{51,102,0}
\definecolor{myred}{RGB}{204,0,0}

\newcommand{\transparentlightblue}{\cellcolor{lightblue!60}}

\begin{table*}[h!]
\centering
\setlength{\tabcolsep}{5pt}
\renewcommand{\arraystretch}{1.12}
\caption{\textbf{Evaluation on STI-Bench.} \colorbox{orange}{Orange} = best and \colorbox{lightorange}{Light Orange} = second best. Note: The random baseline for all tasks is 20\% and @30f denotes evaluated with 30 frames.}
\resizebox{\textwidth}{!}{
\begin{tabular}{lcccccccccc}
\toprule
\textbf{Models} & \textbf{Rank} & \textbf{Avg.}
& \multicolumn{3}{c}{\textbf{Static Understanding}}
& \multicolumn{5}{c}{\textbf{Dynamic Understanding}} \\
\cmidrule(lr){4-6} \cmidrule(lr){7-11}
& & &
\makecell{\textbf{Dim.}\\\textbf{Meas.}} &
\makecell{\textbf{Spatial}\\\textbf{Relation}} &
\makecell{\textbf{3D Video}\\\textbf{Grounding}} &
\makecell{\textbf{Disp.}\\\textbf{\& P.L.}} &
\makecell{\textbf{Speed}\\\textbf{\& Acc.}} &
\makecell{\textbf{Ego}\\\textbf{Orient.}} &
\makecell{\textbf{Traj.}\\\textbf{Desc.}} &
\makecell{\textbf{Pose}\\\textbf{Est.}} \\
\midrule

\rowcolor{lightblue!20}
\multicolumn{11}{l}{\textit{Proprietary Models (API)}} \\
GPT-4o & 10 & 34.8 & 27.1 & 51.8 & 29.0 & 23.2 & 35.4 & 33.7 & 32.0 & 53.6 \\
Gemini-2.0-Flash & 6 & 38.7 & 31.9 & 50.0 & 31.8 & 27.7 & 32.1 & 10.8 & 38.5 & \cellcolor{lightorange}61.3 \\
Claude-3.7-Sonnet & 3 & 40.5 & 29.8 & 45.5 & 35.7 & \cellcolor{lightorange}28.9 & \cellcolor{orange}38.8 & 40.0 & 47.4 & \cellcolor{orange}62.6 \\
Gemini-2.5-Pro & 1 & \cellcolor{orange}41.4 & \cellcolor{orange}38.7 & \cellcolor{orange}53.8 & 36.9 & \cellcolor{orange}33.9 & 33.1 & \cellcolor{orange}52.5 & 47.4 & 50.4 \\

\midrule
\rowcolor{lightblue!20}
\multicolumn{11}{l}{\textit{Open-source Models}} \\
MiniCPM-V-2.6 & 13 & 26.9 & 27.7 & 44.5 & 29.0 & 19.0 & 25.7 & 7.0 & 30.8 & 35.6 \\
VideoChat-R1 & 11 & 32.8 & 23.2 & 47.3 & 31.5 & 22.4 & 31.1 & 26.0 & 47.9 & 48.3 \\
VideoLLaMA3-7B & 9 & 35.2 & 29.4 & 48.6 & 36.1 & 21.5 & 36.7 & 23.2 & \cellcolor{orange}54.6 & 48.1 \\
VideoChat-Flash & 8 & 36.3 & \cellcolor{lightorange}33.6 & 51.4 & 33.1 & 27.1 & 32.3 & 22.2 & \cellcolor{lightorange}54.2 & 51.4 \\
InternVL2.5-78B & 7 & 38.5 & 29.9 & \cellcolor{lightorange}52.8 & 31.6 & 24.9 & 37.2 & \cellcolor{lightorange}49.2 & 43.6 & 53.6 \\
Qwen2.5-VL-72B & 2 & \cellcolor{lightorange}40.7 & 31.5 & 47.6 & \cellcolor{lightorange}39.1 & 25.1 & \cellcolor{lightorange}38.4 & 43.8 & 51.3 & 60.6 \\
GeoThinker@30f&12&27.6&15.2&47.9&30.2&23.4&17.8&15.1&34.6&36.9 \\
\midrule
\rowcolor{lightblue!20}
\multicolumn{11}{l}{\textit{Ours}} \\
\name@16f&4&40.1&24.2&47.2&\cellcolor{orange}41.9&25.7&33.8&41.6&51.2&55.2\\
\name@30f&5&39.9&22.1&48.6&38.1&23.1&33.5&45.9&51.2&56.6\\

\bottomrule
\end{tabular}
}
\label{tab:appendix_stibench}
\end{table*}

Table~\ref{tab:appendix_stibench} evaluates \name with different input frame budgets. Under the same 30-frame evaluation setting, \name improves the overall score from GeoThinker's 27.6 to 39.9. The largest gains occur in dynamic understanding, including ego orientation (+30.8 points), pose estimation (+19.7), and trajectory description (+16.6). With 16 evaluation frames, \name achieves an overall score of 40.1, ranking fourth among the compared configurations, and obtains the highest reported 3D video grounding score of 41.9. These results demonstrate substantial improvements in spatiotemporal understanding that persist across evaluation frame budgets.

\subsection{Comparison to GeoThinker}
\label{sec:appendix_geothinker_comparison}

\begin{figure}[h]
  \centering
    \includegraphics[width=\textwidth]{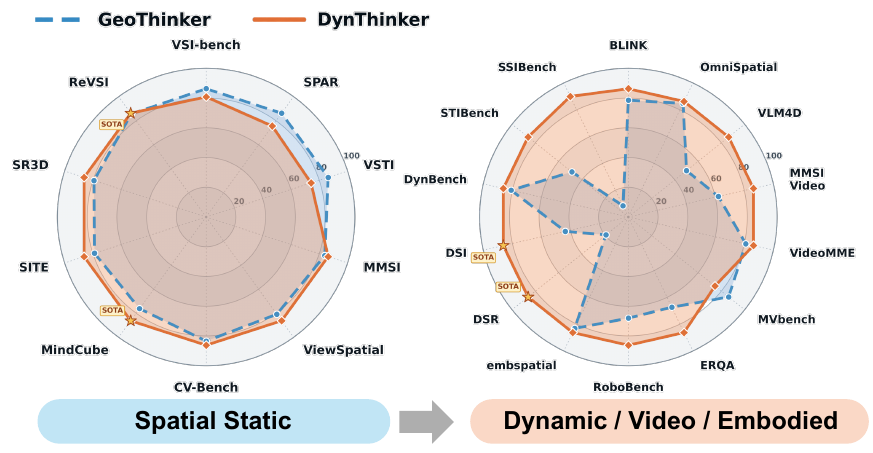} 
    \caption{
    \textbf{Comparison between GeoThinker and \name across static spatial, dynamic, video understanding, and embodied benchmarks.} On static spatial tasks (left), \name improves on several benchmarks while obtaining competitive scores on VSI-Bench, SPAR, and VSTI. On dynamic, video, and embodied tasks (right), \name improves broadly, with MVBench as the main exception.
    }
    \label{fig:appendix_geothinker_comparison}
\end{figure}

Figure \ref{fig:appendix_geothinker_comparison} provides a broader view of how \name changes the capability profile of its GeoThinker baseline. On static spatial benchmarks, the results are mixed: \name improves on ReVSI, MindCube, SITE, SR3D, and several other tasks, while GeoThinker retains an advantage on VSI-Bench, SPAR, and VSTI. The improvement is more consistent on dynamic reasoning, video understanding, and embodied benchmarks. \name shows particularly large gains on DSR, DSI, and VLM4D, and also improves across most of the remaining benchmarks in the right panel. MVBench is the notable exception, where GeoThinker performs better. Overall, the comparison shows that \name’s strongest advance over GeoThinker lies in dynamic and embodied understanding, alongside a mixed but generally competitive static spatial profile.

\clearpage
\subsection{Visualization}
\label{sec:appendix_visualization}

\begin{figure}[h]
  \centering
    \includegraphics[width=\textwidth]{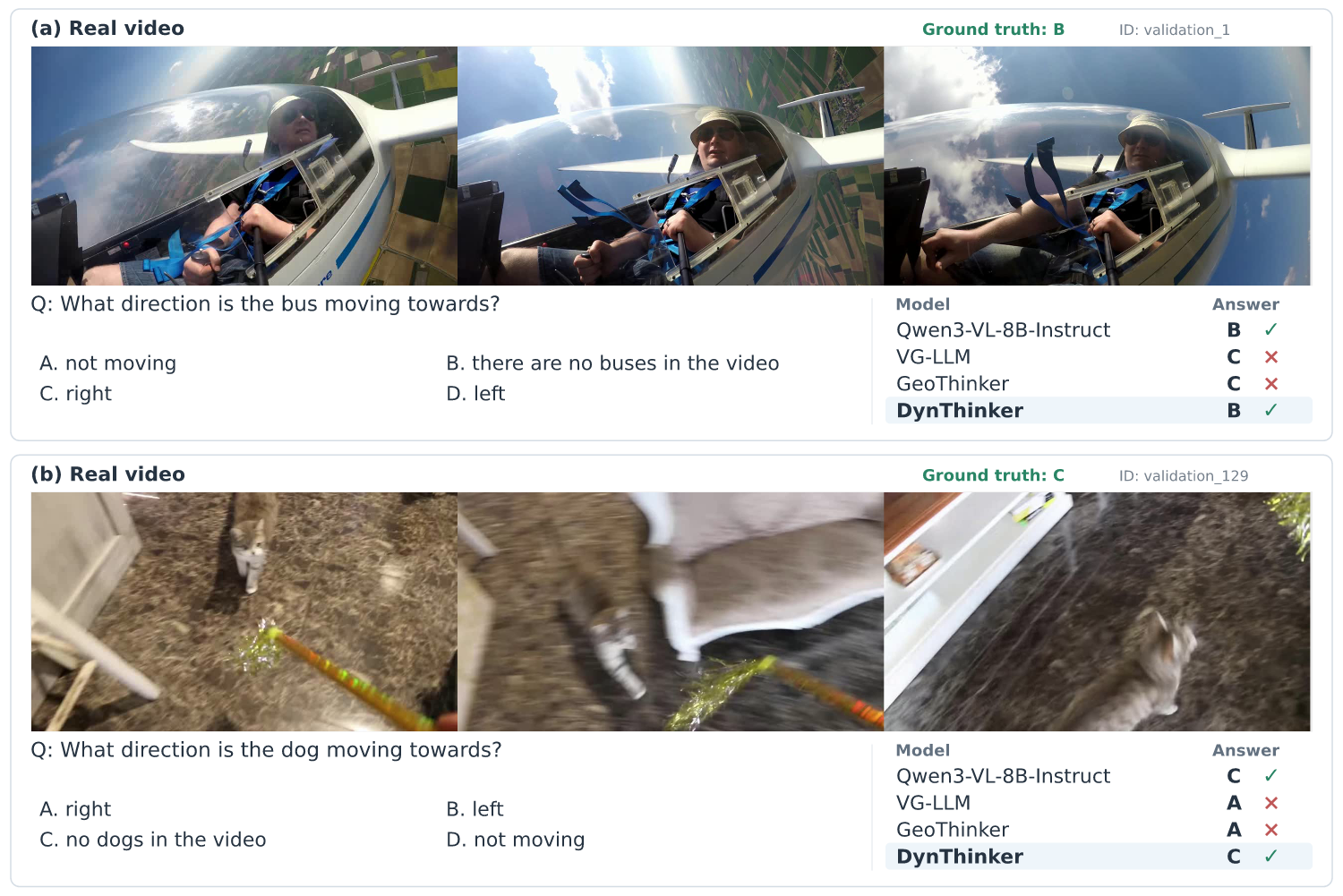} 
    \caption{\textbf{Physical false-premise rejection on real videos.}
\name correctly identifies absent objects, preserving the pretrained model's grounded responses where spatially specialized models hallucinate motion directions.}
    \label{fig:false_premise_real}
\end{figure}

\begin{figure}[h]
  \centering
    \includegraphics[width=\textwidth]{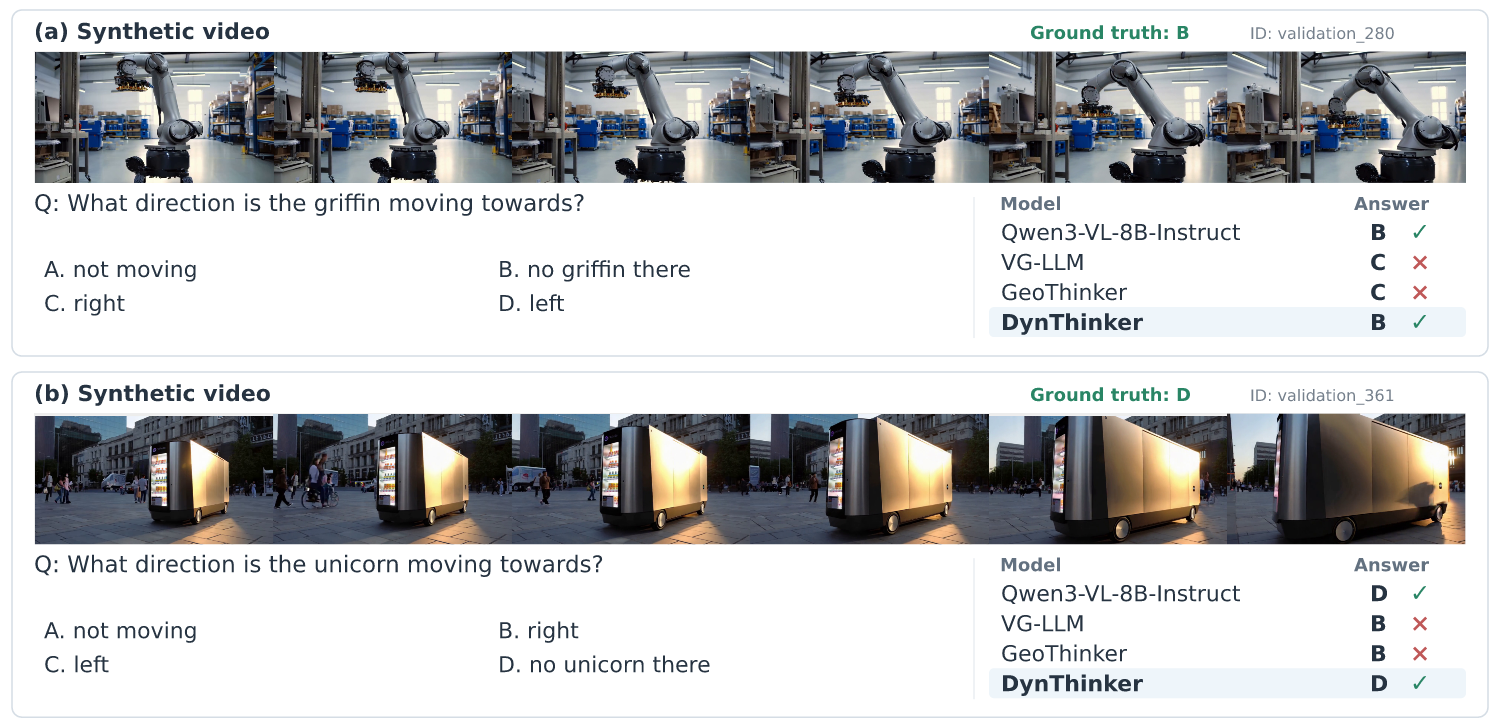} 
    \caption{\textbf{Physical false-premise rejection on synthetic videos.}
\name correctly rejects questions about nonexistent griffins and unicorns, while VG-LLM and GeoThinker assign unsupported motion directions.}
    \label{fig:false_premise_synthetic}
\end{figure}

\begin{figure}[h]
  \centering
    \includegraphics[width=\textwidth]{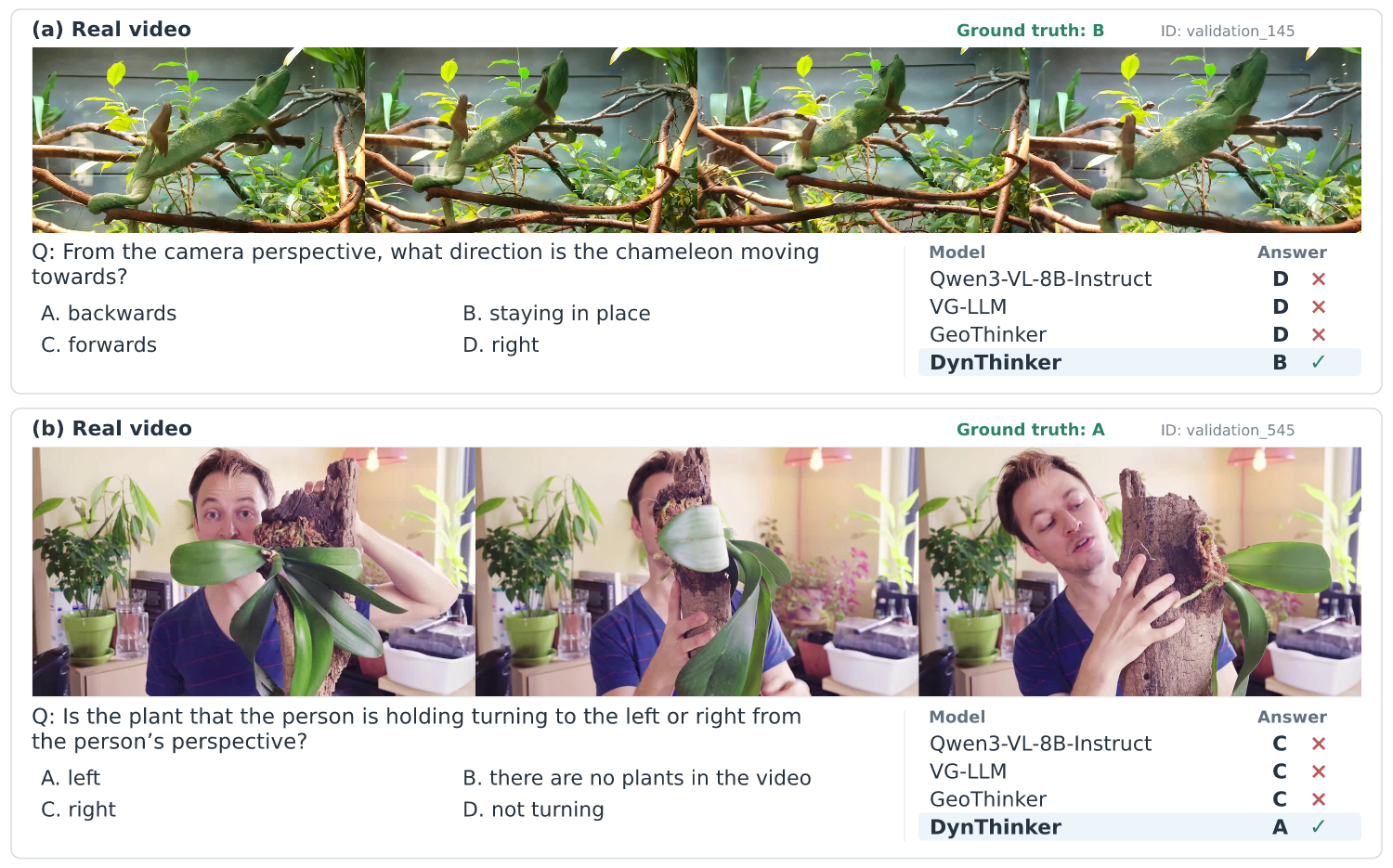} 
    \caption{\textbf{Motion and rotation understanding on real VLM4D videos.}
\name correctly answers stationary-object and perspective-dependent rotation questions that are missed by both the pretrained and spatially specialized models.}
    \label{fig:vlm4d_real_cases}
\end{figure}

\begin{figure}[h]
  \centering
    \includegraphics[width=\textwidth]{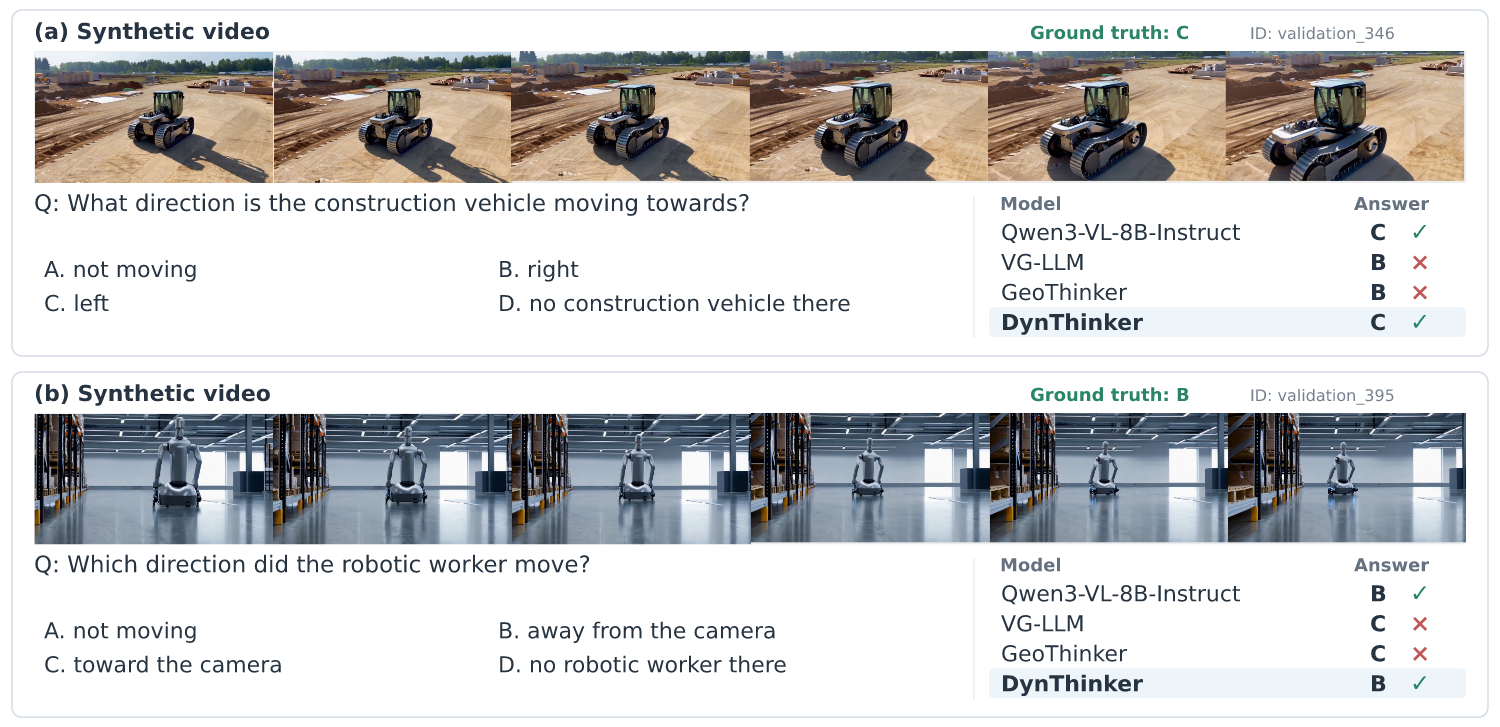} 
    \caption{\textbf{Motion understanding on synthetic VLM4D videos.}
\name preserves correct pretrained predictions for lateral motion and motion away from the camera, where spatially specialized models predict opposite directions.}
    \label{fig:vlm4d_synthetic_cases}
\end{figure}

\clearpage
\subsubsection{Physical grounding and false-premise rejection.}
Figures~\ref{fig:false_premise_real} and~\ref{fig:false_premise_synthetic} illustrate grounding failures examples associated with 3D-focused specialization. Although the queried objects are absent, VG-LLM and GeoThinker accept the false premises and still predict motion directions. In contrast, the pretrained Qwen3-VL-8B model correctly rejects these premises, and \name preserves this behavior across both real and synthetic videos. These qualitative examples support our diagnosis of spatial--dynamic interference: 3D-focused adaptation can weaken reliance on visual evidence when making dynamic judgments.

\subsubsection{Motion and rotation understanding.}
Figures~\ref{fig:vlm4d_real_cases} and~\ref{fig:vlm4d_synthetic_cases} illustrate both the retention examples of pretrained dynamic competence and further improvements in dynamic reasoning. In the real videos, \name correctly identifies the stationary chameleon and the plant's rotation from the person's perspective, correcting errors made by both the pretrained backbone and spatially specialized models. In the synthetic videos, it retains the backbone's correct judgments of vehicle motion and movement away from the camera, whereas VG-LLM and GeoThinker reverse the corresponding directions. Together, these cases show that \name preserves reliable dynamic judgments from pretraining while improving cases that the pretrained model fails to resolve.

\section{Additional results (4D data-scaling)}
\subsection{Further VLM4D comparison}
\label{sec:appendix_vlm4d_comparison}

\begin{table}[h!]
  \centering
  \caption{\textbf{Comparison of official and deterministic evaluation protocols on VLM4D.} The upper block reproduces results reported by MLLM4D, whose official evaluation uses an auxiliary language model to extract the predicted option from model-generated responses. The lower block reports results obtained with our deterministic protocol, which directly matches the output against the valid options A--D without an auxiliary evaluator. To isolate answer parsing from differences in inference frameworks or evaluation prompts, rows marked with $*$ are rescored directly from the model responses released in the official MLLM4D repository rather than regenerated using our framework. \emph{Empty} denotes responses containing no predicted answer, while \emph{Valid} denotes outputs that directly match one of the four option labels. These two columns evaluate output format rather than answer correctness. Real and synthetic averages, as well as the overall score, are weighted by their corresponding sample counts. $\dagger$ denotes the final DynThinker checkpoint evaluated without token compression, and $\ddagger$ denotes the best DynThinker checkpoint trained under the MLLM4D data-scaling setting. Dashes indicate unavailable results.
  }
  \label{tab:appendix_vlm4d_comparison}
  \setlength{\tabcolsep}{3pt}
  \begin{tabular}{l | c  c | c c | c}
    \toprule
    \multirow{2}{*}{\textbf{Model}} & \multirow{2}{*}{\textbf{Real(\%)}} & \multirow{2}{*}{\textbf{Synthetic(\%)}} & \multicolumn{2}{c|}{\textbf{Format(\%)}} & \multirow{2}{*}{\textbf{Overall}}\\

    \cmidrule{4-5}

    &  &  & Empty & Valid & \\

    \midrule
    \rowcolor{black!11}
    \multicolumn{6}{l}{\textbf{MLLM4D Official}} \\
    Qwen3-VL-8B  & 52.1 & 52.4 & -- & -- & 52.2 \\
    MLLM4D\textsubscript{Qwen2.5-VL-7B} & 59.4 & 49.7 & -- & -- & 57.0 \\
    MLLM4D\textsubscript{Qwen3-VL-8B} & 63.1 & 54.4 & -- & -- & 61.0 \\
    MLLM4D-SFT & 59.7 & -- & -- & -- & -- \\
    MLLM4D-RFT & 63.1 & -- & -- & -- & -- \\

    \midrule
    \rowcolor{black!11}
    \multicolumn{6}{l}{\textbf{Ours protocol}} \\

    Qwen3-VL-8B & 58.92 &  63.59 & 0 & 100 & 60.07 \\

    MLLM4D-SFT* &  58.64 &  53.48 & 0.83 & 99.17 & 57.38 \\

    MLLM4D-RFT* &  60.17 & 54.38 & 3.91 & 96.09 & 58.76 \\

    \midrule

    \name\textsubscript{Qwen3-VL-8B} & 61.52 & 59.77 & 0 & 100 & 60.07 \\

    \name\textsubscript{Qwen3-VL-8B} $\dagger$ & 63.54 & 61.79 & 0 & 100 & 62.27 \\
    
    \name\textsubscript{Qwen3-VL-8B} $\ddagger$ & 64.55 & 73.48 & 0 & 100 & 66.74 \\

    \bottomrule
  \end{tabular}
\end{table}

Table~\ref{tab:appendix_vlm4d_comparison} examines whether \name’s VLM4D gains can be attributed to differences in answer extraction. Because rerunning MLLM4D within a different framework may introduce changes in prompts or inference behavior, we directly rescore its officially released model responses using our deterministic matching protocol. The released SFT and RFT predictions already follow the required output format in 99.17\% and 96.09\% of cases, respectively. The remaining failures are all empty responses rather than ambiguous answers requiring semantic extraction. Re-evaluation produces scores comparable to, and generally no higher than, the officially reported results, indicating that our protocol is not more permissive and does not artificially inflate performance. Under the same deterministic evaluation, the final \name model obtains 60.07, which increases to 62.27 when token compression is disabled at inference. The model trained under the MLLM4D data-scaling setting further reaches 66.74, exceeding the best officially reported overall score of 61.0 by 5.74 points. These results confirm that \name’s improvement persists under a fixed, evaluator-free protocol and is not an artifact of auxiliary answer extraction.

\section{Additional results (3D vanilla regime)}
\subsection{Spatial Reasoning}
\label{sec:appendix_spatial_reasoning}

\subsubsection{Setting} 
\textbf{Baseline.} GeoThinker~\citep{li2026thinking} proposes active geometry integration to actively select geometry features from VGGT~\citep{wang2025vggt} for MLLMs, serving as our primary baseline in vanilla regime.

\textbf{Evaluation protocol:} Following the VG-LLM~\citep{zheng2025learning(vgllm)} configuration, we utilize sampled subsets from SPAR-7M \citep{zhang2025flatland(spar)} and the LLaVA-Hound split of LLaVA-Video-178K \citep{zhang2024video} for fine-tuning.We uniformly sample 8 frames per scene for consistency with the baseline.

\subsubsection{Vanilla Regime}

\begin{table*}[h]
  \caption{
  \textbf{Performance comparisons on VSI-Bench (Vanilla regime).} \name matches or outperforms GeoThinker and other baseline across different model scales, achieving the best overall performance with the Qwen3-VL-8B variant.
  }
  \vspace{-1mm}
  \label{tab:appendix_vsi_bench}
  \begin{center}
    \begin{small}
      \begin{sc}
        \resizebox{\textwidth}{!}{%
        \begin{tabular}{l|c|cccc|cccc}
          \toprule
          \multirow{2}{*}{Methods} & \multirow{2}{*}{Avg.}
          & \multicolumn{4}{c|}{\cellcolor{orange!10}Numerical Answer}
          & \multicolumn{4}{c}{\cellcolor{yellow!10}Multiple-Choice Answer} \\
          \cmidrule(lr){3-6}\cmidrule(lr){7-10}
          & &  Obj.Count & Abs. Dist. & Obj. Size & Room Size & Rel. Dist. & Rel. Dir. & Route Plan & Appr. Order  \\
          \midrule
          \rowcolor{black!11}
          \multicolumn{10}{l}{\textbf{Baseline}} \\
          Chance Level (Random)  & --   & --   & --   & --   & --   & 25.0   & 36.1   & 28.3   & 25.0 \\
          Chance Level (Frequency) & 34.0 & 62.1 & 32.0 & 29.9 & 33.1 & 25.1 & 47.9 & 28.4 & 25.2 \\
          \midrule
          \rowcolor{black!8}
          \multicolumn{10}{l}
          {\textbf{Proprietary Models (API)}} \\
          GPT-4o\citep{hurst2024gpt4o}            & 34.0 & 46.2 & 5.3  & 43.8 & 38.2 & 37.0 & 41.3 & 31.5 & 28.5 \\
          Gemini-1.5 Flash\citep{team2024gemini(gemini1.5)}   & 42.1 & 49.8 & 30.8 & 53.5 & 54.4 & 37.7 & 41.0 & 31.5 & 37.8 \\
          Gemini-1.5 Pro\citep{team2024gemini(gemini1.5)}     & 45.4 & 56.2 & 30.9 & 64.1 & 43.6 & 51.3 & 46.3 & 36.0 & 34.6 \\
          \midrule
          \rowcolor{black!11}
          \multicolumn{10}{l}{\textbf{Open-sourced Models}} \\
          LLaVA-OneVision-7B\citep{li2024llava-onevision}      & 32.4 & 47.7 & 20.2 & 47.4 & 12.3 & 42.5 & 35.2 & 29.4 & 24.4 \\
          LLaVA-OneVision-72B\citep{li2024llava-onevision}     & 40.2 & 43.5 & 23.9 & 57.6 & 37.5 & 42.5 & 39.9 & 32.5 & 44.6 \\
          LLaVA-NeXT-Video-7B\citep{liu2024llavanext}     & 35.6 & 48.5 & 14.0 & 47.8 & 24.2 & 43.5 & 42.4 & 34.0 & 30.6 \\
          LLaVA-NeXT-Video-72B\citep{liu2024llavanext}  & 40.9 & 48.9 & 22.8 & 57.4 & 35.3 & 42.4 & 36.7 & 35.0 & 48.6 \\
          InternVL2-8B\citep{chen2024internvl2}           & 34.6 & 23.1 & 28.7 & 48.2 & 39.8 & 36.7 & 30.7 & 29.9 & 39.6 \\
          InternVL2-40B\citep{chen2024internvl2}         & 36.0 & 34.9 & 26.9 & 46.5 & 31.8 & 42.1 & 32.2 & 34.0 & 39.6 \\

          Qwen2.5-VL-3B\citep{Qwen2.5-VL} & 28.6 & 32.7 & 19.5 & 17.3 & 25.1 & 37.3 & 44.9 & 30.4 & 21.8 \\
          Qwen2.5-VL-7B\citep{Qwen2.5-VL} & 29.3 & 25.2 & 10.5 & 36.4 & 29.6 & 38.4 & 38.0 & 29.8 & 26.8 \\

          \midrule
          \rowcolor{black!11}
          \multicolumn{10}{l}{\textbf{Open-source Spatial Intelligence Models}} \\

          SPAR-8B\citep{zhang2025flatland(spar)} & 44.1 & -- & -- & -- & -- & -- & -- & -- & -- \\

          SpatialLadder-3B\citep{li2025spatialladder} & 44.8 & -- & -- & -- & -- & -- & -- & -- & -- \\
          
          Spatial-MLLM-4B\citep{wu2025spatial(spatialmllm)}  & 48.4 & 65.3 & 34.8 & 63.1 & 45.1 & 41.3 & 46.2 & 33.5 & 46.3 \\
        
          VG-LLM-4B\citep{zheng2025learning(vgllm)} & 46.7 & 67.6 & 37.6 & 55.2 & 52.5 & 48.0 & 44.7 & 31.9 & 35.5 \\

          VG-LLM-8B~\citep{zheng2025learning(vgllm)}  & 49.7 & 68.1 & 38.7 & 59.0 & 61.1 &45.5 &44.9 &26.8 &53.4\\

          GeoThinker\textsubscript{Qwen2.5-VL-3B}~\citep{li2026thinking}  & 48.9 & 68.5 & 36.1 & 57.3 & 62.5 & 43.7 & 47.9 & 34.5 & 40.9 \\
          
          GeoThinker\textsubscript{Qwen2.5-VL-7B}~\citep{li2026thinking}  & 50.5 & 69.5 & 38.5 & 57.9 & 62.2 & 45.2 & 46.2 & 31.4 & 52.6 \\

          GeoThinker\textsubscript{Qwen3-VL-4B}~\citep{li2026thinking}  & 51.3 & 67.4 & 41.9 & 64.0 & 61.1 & 50.5 & 45.0 & 31.9 & 48.3 \\
          
          GeoThinker\textsubscript{Qwen3-VL-8B}~\citep{li2026thinking}  & 52.4 & 68.0 & 44.0 & 60.9 & 61.7 & 53.5 & 47.9 & 31.9 & 51.1\\

        \midrule
          \rowcolor{navyblue!11}
          \multicolumn{10}{l}{\textbf{Ours}} \\

          \name\textsubscript{Qwen2.5-VL-7B} & 50.5 & 67.6 & 36.3 & 57.7 & 62.1 & 46.3 & 43.1 & 33.5 & 57.4 \\

          \name\textsubscript{Qwen3-VL-4B} & 56.5 & 70.0 & 48.2 & 75.4 & 58.2 & 57.3 & 49.2 & 33.4 & 57.4 \\

          \name\textsubscript{Qwen3-VL-8B} & \sota{58.1} & 67.7 & 50.5 & 74.7 & 58.2 & 60.2 & 53.9 & 35.0 & 64.7 \\

          \bottomrule
        \end{tabular}%
        }
      \end{sc}
    \end{small}
  \end{center}
  \vskip -0.2in
\end{table*}

Table~\ref{tab:appendix_vsi_bench} reports VSI-Bench results under the vanilla training regime. With Qwen3-VL-4B and Qwen3-VL-8B, \name achieves overall scores of 56.5 and 58.1, improving over the corresponding GeoThinker variants by 5.2 and 5.7 points, respectively. The 4B variant already exceeds all listed baselines, while the 8B variant achieves the highest overall score in this comparison.

The improvements cover several geometric and relational reasoning tasks. On the 4B backbone, object size estimation, relative distance reasoning, and approach ordering improve by 11.4, 6.8, and 9.1 points over GeoThinker, respectively. On Qwen2.5-VL-7B, \name maintains GeoThinker's overall score of 50.5 while improving route planning and approach ordering. These results demonstrate strong spatial performance with Qwen3-VL series backbone and extend the benefits observed in the main experiments to the vanilla regime.

\subsection{D-EMA Ablation study}
\label{sec:appendix_edma_ablation}

\subsubsection{Group-wise Target Drift Budget}
We specify the preservation strength of each pretrained parameter group
$k$ through a target drift budget
\begin{equation}
\delta_k = 1-\rho_k^{N_{\mathrm{ref}}},
\end{equation}
where $N_{\mathrm{ref}}$ is the number of D-EMA updates in the reference
training schedule. With the shadow parameters initialized at
$\bar{\theta}_0^{(k)}=\theta_0^{(k)}$, unrolling the group-wise update gives
\begin{equation}
\bar{\theta}_{N_{\mathrm{ref}}}^{(k)}
=
\rho_k^{N_{\mathrm{ref}}}\theta_0^{(k)}
+
(1-\rho_k)
\sum_{t=1}^{N_{\mathrm{ref}}}
\rho_k^{N_{\mathrm{ref}}-t}\theta_t^{(k)}.
\end{equation}
Thus, $1-\delta_k$ is the explicit coefficient on the pretrained
initialization, while $\delta_k$ is the total coefficient assigned to
online parameter states encountered during post-training. A smaller
$\delta_k$ places greater weight on the initialization. A larger
$\delta_k$ places greater weight on the adapted trajectory. This budget
controls the aggregation weights rather than directly constraining the
drift of the online parameters. At the two extremes, $\delta_k=0$
retains the initialization, whereas $\delta_k=1$ selects the final
online parameters.
Given $\delta_k$, the decay for the reference schedule is
\begin{equation}
\rho_k=(1-\delta_k)^{1/N_{\mathrm{ref}}}.
\end{equation}
For a target schedule with $N_{\mathrm{target}}$ D-EMA updates, we use
\begin{equation}
\rho_k'=(1-\delta_k)^{1/N_{\mathrm{target}}},
\end{equation}
which preserves the same target budget across training horizons.

\subsubsection{D-EMA drift allocation.}
\begin{table}[h]
\centering
\caption{
\textbf{Ablation of D-EMA drift allocation.}
All post-trained models are trained under the spatial-reasoning vanilla regime with SDF enabled.
We vary the target drift budgets for MLP layers $j>4$ and other pretrained parameters.
Newly introduced SDF parameters are excluded from averaging.
($\cdot$) denotes the performance change compared to pre-train model.
}
\label{tab:ema_drift_allocation}
\vspace{-1mm}
\small
\resizebox{0.7\columnwidth}{!}{%
\begin{tabular}{lcccc}
\toprule
\multirow{2}{*}{Strategy}
& \multicolumn{2}{c}{Target Drift Budget}
& \multirow{2}{*}{VSI-Bench}
& \multirow{2}{*}{VLM4D} \\
\cmidrule(lr){2-3}
& MLP ($j>4$) & Others & & \\
\midrule

\rowcolor{black!11}
\multicolumn{5}{l}{\textbf{Qwen2.5-VL-7B}} \\
Pretrain & -- & -- & 29.34 & 43.80 \\
\midrule
\multirow{3}{*}{D-EMA} & 12.5\% & 12.5\% & 34.44\textcolor{blue}{(+5.10)} & \textbf{55.17}\textcolor{blue}{(+11.37)} \\
 & 100\%  & 100\%  & 49.65\textcolor{blue}{(+20.31)} & 46.64\textcolor{blue}{(+2.84)} \\
 & 12.5\% & 100\%  & \textbf{50.37}\textcolor{blue}{(+21.03)} & 48.56\textcolor{blue}{(+4.76)} \\

\midrule
\rowcolor{black!11}
\multicolumn{5}{l}{\textbf{Qwen3-VL-8B}} \\
Pretrain & -- & -- & 57.75 & 60.07 \\
\midrule
\multirow{11}{*}{D-EMA} & 5\%    & 5\%   &55.73\textcolor{orange}{(-2.02)}&  58.86\textcolor{orange}{(-1.21)}  \\
 & 5\%    & 12.5\%   & 56.50\textcolor{orange}{(-1.25)} & 60.02\textcolor{orange}{(-0.05)} \\
 & 5\%    & 25\%   &57.02\textcolor{orange}{(-0.73)} & 57.81\textcolor{orange}{(-2.26)}  \\
 & 5\%    & 50\%   & 57.17\textcolor{orange}{(-0.58)} & 58.20\textcolor{orange}{(-1.87)} \\

\cmidrule{2-5}
 
 & 12.5\% & 12.5\% & \textbf{58.16}\textcolor{blue}{(+0.41)} & \textbf{61.78}\textcolor{blue}{(+1.71)} \\
 & 12.5\% & 25\%   & 57.51\textcolor{orange}{(-0.24)} & 56.99\textcolor{orange}{(-3.08)} \\
 & 12.5\% & 50\%   & 58.06\textcolor{blue}{(+0.31)} & 56.05\textcolor{orange}{(-4.02)} \\

\cmidrule{2-5}
 
 & 25\% & 25\% & 57.83\textcolor{blue}{(+0.08)} & 57.70\textcolor{orange}{(-2.37)} \\
 & 25\% & 50\% & 57.64\textcolor{blue}{(-0.11)} & 55.39\textcolor{orange}{(-4.68)} \\

\cmidrule{2-5}
 
 & 50\%   & 50\%   & 57.09\textcolor{orange}{(-0.66)} & 52.80\textcolor{orange}{(-7.27)} \\

\bottomrule
\end{tabular}%
}
\end{table}
Table~\ref{tab:ema_drift_allocation} examines how the drift budget should be allocated across pretrained parameter groups. For Qwen2.5-VL-7B, applying a shared 12.5\% budget favors dynamic retention but substantially restricts spatial adaptation, yielding only 34.44 on VSI-Bench. Assigning a 12.5\% budget to MLPs in blocks $j>4$ and a 100\% budget to the remaining pretrained parameters improves VSI-Bench to 50.37 while retaining 48.56 on VLM4D, demonstrating the benefit of asymmetric consolidation. For Qwen3-VL-8B, the shared 12.5\% configuration performs best among the evaluated allocations, reaching 58.16 on VSI-Bench and 61.78 on VLM4D. We therefore use budgets of 12.5\%/100\% for Qwen2.5-VL-7B and a shared 12.5\% budget for the Qwen3-VL models.

\subsubsection{Training and component ablations.}
\begin{table}[h]
\centering
\caption{
\textbf{Training and component ablations under the Static 3D-reasoning vanilla regime.}
We compare pretrained backbones, Static 3D SFT, EMA-based preservation, and SDF integration.
For Qwen2.5-VL-7B, D-EMA uses target drift budgets of 12.5\% for MLP layers $j>4$ and 100\% for other pretrained parameters.
For Qwen3-VL, D-EMA uses a shared budget of 12.5\%.
}
\label{tab:ema_ablation}
\vspace{-1mm}
\small
\resizebox{0.7\columnwidth}{!}{%
\begin{tabular}{lcc}
\toprule
Training Configuration & VSI-Bench & VLM4D \\
\midrule

\rowcolor{black!11}
\multicolumn{3}{l}{\textbf{Qwen2.5-VL-7B}} \\
Pretrained                    & 29.34 & 43.80 \\
Static 3D SFT                   & 49.37 & 46.80 \\
Static 3D SFT + D-EMA            & 49.73 & 47.57 \\
Static 3D SFT + D-EMA + SDF      & \textbf{50.37} & \textbf{48.56} \\

\midrule
\rowcolor{black!11}
\multicolumn{3}{l}{\textbf{Qwen3-VL-4B}} \\
Pretrained                    & 54.90 & 57.10 \\
Static 3D SFT                   & 52.61 & 46.75 \\
Static 3D SFT + D-EMA      & 55.04 & 57.37 \\
Static 3D SFT + D-EMA + SDF
                              & \textbf{56.56} & \textbf{59.74} \\

\midrule
\rowcolor{black!11}
\multicolumn{3}{l}{\textbf{Qwen3-VL-8B}} \\
Pretrained                    & 57.75 & 60.07 \\
Static 3D SFT                   & \textbf{58.16} & 59.74 \\
Static 3D SFT + D-EMA      & 58.07 & 59.74 \\
Static 3D SFT + D-EMA + SDF
                              & \textbf{58.16} & \textbf{61.78} \\

\bottomrule
\end{tabular}%
}
\end{table}
Table~\ref{tab:ema_ablation} isolates the effects of Static 3D SFT, D-EMA, and SDF across three backbones. On Qwen3-VL-4B, direct Static 3D SFT reduces VSI-Bench from 54.90 to 52.61 and VLM4D from 57.10 to 46.75. D-EMA recovers both capabilities to 55.04 and 57.37, while adding SDF further improves them to 56.56 and 59.74. On Qwen3-VL-8B, Static 3D SFT causes only minor dynamic degradation, leaving limited room for D-EMA to improve the standalone scores, whereas SDF raises VLM4D from 59.74 to 61.78 while maintaining Static 3D performance. We hypothesize that this weaker degradation arises because Qwen3-VL already acquires substantial Static 3D knowledge during pretraining, with the larger 8B backbone retaining these priors more robustly during specialization. Nevertheless, the 52.4 VSI-Bench score of GeoThinker with the same Qwen3-VL-8B backbone in Table~\ref{tab:appendix_vsi_bench} suggests that jointly optimizing newly introduced modules can still interfere with established representations. Thus, although D-EMA provides little standalone improvement for Qwen3-VL-8B, it serves as a consolidation mechanism during joint adaptation, stabilizing pretrained representations while allowing the newly introduced SDF parameters to remain fully plastic. Overall, D-EMA is most beneficial when specialization causes substantial capability degradation, while also safeguarding pretrained representations during perceptual expansion.

\section{Additional results (4D vanilla regime)}
\subsection{Temporal Dynamics}
\label{sec:appendix_temporal_dynamics}
\subsubsection{Setting} 
\textbf{Evaluation protocol:} We construct a controlled \emph{temporal-dynamic vanilla regime} by augmenting the spatial reasoning vanilla training mixture with 10\% of the MLLM4D~\citep{yin2026mllm4d} training data (200k). The spatial mixture consists of sampled subsets of SPAR-7M~\citep{zhang2025flatland(spar)} (234k) and the LLaVA-Hound split of LLaVA-Video-178K~\citep{zhang2024video} (64k). All models are trained with 8 input frames. We evaluate spatial reasoning on VSI-Bench and temporal-dynamic understanding on VLM4D, comparing alternative capability-preservation strategies and examining inference-time frame scaling.

\subsubsection{Preservation strategies} 

\begin{table}[h!]
  \centering
  \caption{\textbf{Comparison of capability-preservation strategies under the temporal-dynamic vanilla regime.} We compare MLP freezing, weight pulling, spectral regularization, conventional EMA, and D-EMA using the same training mixture and 8-frame training inputs. All results use Qwen3-VL-4B unless otherwise specified. Interval step denotes the number of optimization steps between applications of the corresponding strategy.}
\label{tab:appendix_preservation_comparison}
  \setlength{\tabcolsep}{3pt}
  \begin{tabular}{l | c |c c}
    \toprule
    \textbf{Strategy} & \textbf{Update interval} & \textbf{VSI-Bench} & \textbf{VLM4D} \\
    \midrule
    \rowcolor{black!11}
    \multicolumn{4}{l}{\textbf{Baseline}} \\

    Pre-train Model & -- & 54.90 & 57.10 \\
    Direct SFT & -- & 52.61 & 46.75 \\
    
    \midrule
    \rowcolor{black!11}
    \multicolumn{4}{l}{\textbf{Frozen MLP}} \\

    Frozen All MLP & -- & 51.41 & 50.27 \\

    \midrule
    \rowcolor{black!11}
    \multicolumn{4}{l}{\textbf{Weight Pulling}} \\

    weight ratio 0.05 & 100 & 50.78 & 50.99  \\
    weight ratio 0.1 & 100 & 49.90 & 49.50  \\
    weight ratio 0.2 & 100 & 48.07 &  48.56 \\
    weight ratio 0.75 & 100 & 47.97 & 47.90  \\

    \midrule
    \rowcolor{black!11}
    \multicolumn{4}{l}{\textbf{Spectral regularization}} \\

    spectral weight 1e9 & 2 & 49.89 & 50.66  \\
    spectral weight 5e9 & 2 & 51.49 & 51.37  \\

    \midrule
    \rowcolor{black!11}
    \multicolumn{4}{l}{\textbf{EMA}} \\

    EMA decay 0.999 & 10 & 52.55 & 55.12 \\

    \midrule
    \rowcolor{navyblue!11}
    \multicolumn{4}{l}{\textbf{D-EMA}\textit{(Ours)}} \\

    \name  & 10 & 55.31 & 60.57 \\
    \name\textsubscript{Qwen3-VL-8B}  & 10 & 58.48 & 64.42 \\

    \bottomrule
  \end{tabular}
\end{table}

Under the matched Qwen3-VL-4B setting in Table~\ref{tab:appendix_preservation_comparison}, direct SFT scores 52.61 on VSI-Bench and 46.75 on VLM4D. Freezing all MLPs raises VLM4D to 50.27 but lowers VSI-Bench to 51.41. Weight pulling periodically moves the trained weights toward their pretrained values, whereas spectral regularization penalizes changes in the spectra of weight matrices. Its large coefficients compensate for the small numerical scale of the penalty. Both strategies improve VLM4D over direct SFT but reduce VSI-Bench performance, and stronger weight pulling further degrades both scores. Conventional EMA, which uses a uniform decay, offers a better balance at 52.55 and 55.12, but still falls short of the pretrained VLM4D score of 57.10. In contrast, D-EMA reaches 55.31 on VSI-Bench and 60.57 on VLM4D, exceeding the pretrained scores on both benchmarks and achieving the best balance among the compared strategies. For Qwen3-VL-8B, the reported scores are 58.48 on VSI-Bench and 64.42 on VLM4D.

\subsubsection{Inference-time frame scaling} 
\begin{table}[h]
\vspace{-1mm}
\centering
\small
\setlength{\tabcolsep}{1.5mm}
\caption{\textbf{Inference-time frame ablation under the temporal-dynamic vanilla regime.} All models are trained with 8-frame inputs and evaluated with different frame budgets on VSI-Bench and VLM4D, with and without visual token compression. $\ast$ denotes sparse VDPM decoding with a maximum of 8 sampled frames, as described in Model Configurations. unmarked variants decode all input frames. OOM indicates an out-of-memory failure.}
\label{tab:appendix_temporal_frame_ablation}
\resizebox{0.8\linewidth}{!}{
\begin{tabular}{llccccc}
\toprule
\multirow{2}{*}{\textbf{Model}} & \multirow{2}{*}{\textbf{Benchmark}} & \multicolumn{5}{c}{\textbf{\# Frames}} \\
\cmidrule(lr){3-7}
 &  & \textbf{8} & \textbf{16} & \textbf{32} & \textbf{64} & \textbf{128} \\

\midrule
\rowcolor{black!11}
\multicolumn{7}{l}{\textbf{With token compression}} \\

\multirow{2}{*}{\name\textsubscript{Qwen3-VL-4B}}
& VSI & 47.72 & 49.37 & 53.14 & 56.37 & \sota{57.24} \\
& VLM4D & \sota{57.65} & 53.41 & 53.41 & 53.02 & 51.87 \\

\midrule
\multirow{2}{*}{\name\textsubscript{Qwen3-VL-8B}}
& VSI & 49.93 & 53.52 & 57.11 & \sota{58.87} & OOM \\
& VLM4D & \sota{62.33} & 61.56 & 60.51 & 60.51 & OOM \\

\midrule
\multirow{2}{*}{\name\textsubscript{Qwen3-VL-8B}}*
& VSI & 50.03 & 53.48 & 56.44 & 58.39 & \sota{59.11} \\
& VLM4D & \sota{62.11} & 61.72 & 60.07 & 60.62 & 60.35 \\

\midrule
\rowcolor{black!11}
\multicolumn{7}{l}{\textbf{Without token compression}} \\

\multirow{2}{*}{\name\textsubscript{Qwen3-VL-4B}}
& VSI & 47.72 & 52.26 & 55.27 & \sota{57.18} & 57.16 \\
& VLM4D & 57.65 & \sota{58.53} & 58.48 & 58.09 & 55.61 \\

\midrule
\multirow{2}{*}{\name\textsubscript{Qwen3-VL-8B}}
& VSI & 49.93 & 54.67 & 58.48 & \sota{58.93} & OOM \\
& VLM4D & 62.33 & \sota{64.42} & 62.72 & 63.93 & OOM \\

\midrule
\multirow{2}{*}{\name\textsubscript{Qwen3-VL-8B}}*
& VSI & 50.03 & 54.30 & 57.60 & \sota{58.27} & 57.97 \\
& VLM4D & 62.11 & \sota{64.15} & 62.99 & 63.60 & 62.44 \\
 
\bottomrule
\end{tabular}}
\vspace{-2pt}
\label{tab:comparisons_vsidebiased}
\vspace{-5mm}
\end{table}

Table~\ref{tab:appendix_temporal_frame_ablation} reveals different scaling behaviors for static and dynamic reasoning. Without token compression, the 8B model improves on VSI-Bench from 49.93 at 8 frames to 58.93 at 64 frames, whereas its VLM4D accuracy peaks at 64.42 with 16 frames. Removing token compression consistently improves VLM4D for inputs longer than 8 frames, suggesting that temporal-dynamic reasoning benefits from finer per-frame visual evidence. Capping VDPM decoding at 8 sampled frames further enables the 8B model to process 128-frame inputs without out-of-memory failures, achieving 57.97 on VSI-Bench and 62.44 on VLM4D without token compression.

\section{Additional results (sensitivity analysis)}
\subsection{Selection of the MLP Layer Boundary}
\label{sec:mlp_layer_boundary}

\begin{figure}[h]
  \centering
    \includegraphics[width=\textwidth]{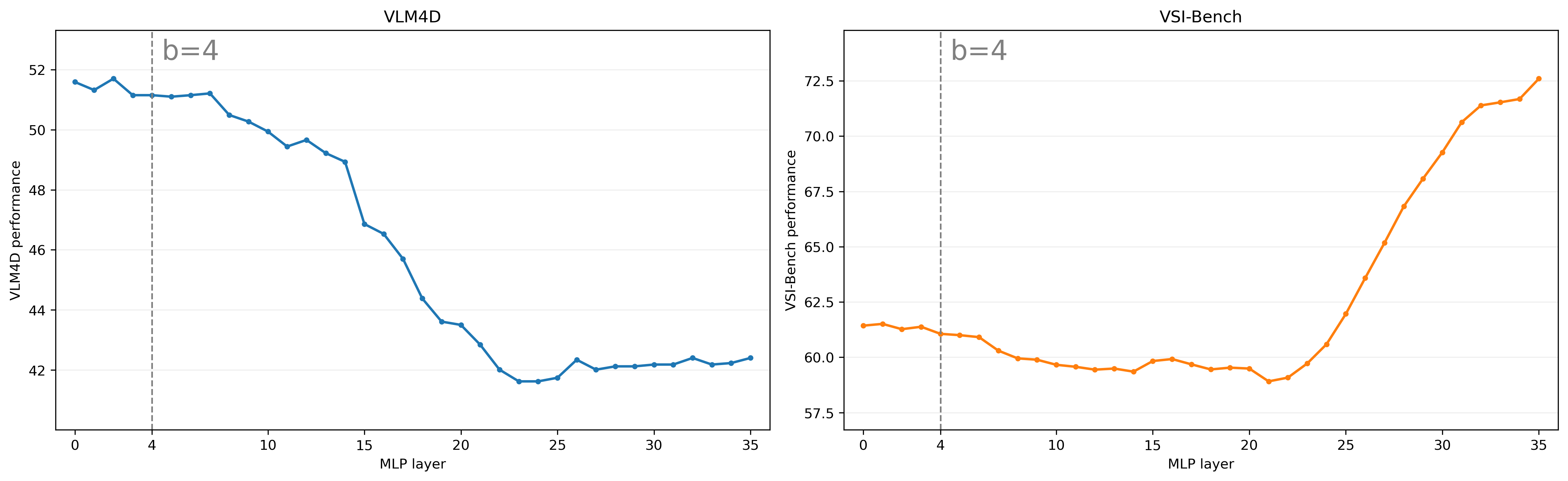} 
    \caption{\textbf{Layer-boundary analysis for selective MLP preservation.} 
    We index the LLM blocks from \(1\) to \(L\), and let \(b\) denote the number of earliest MLP blocks that retain their specialized weights.
    Starting from GeoThinker-Qwen3-VL-8B, we restore MLP weights in layers $j>b$ to their pretrained Qwen3-VL-8B values, while retaining the specialized weights in layers $j\leq b$ and all other parameters. Thus, $b=0$ restores all MLP layers, whereas $b=4$ retains the first four specialized MLP layers. We report VLM4D and VSI-Bench performance as the boundary varies. The dashed line marks our selected boundary $b=4$.}
\label{fig:appendix_mlp_boundary}
\end{figure}
Figure~\ref{fig:appendix_mlp_boundary} examines how the MLP restoration boundary affects dynamic and spatial performance. VLM4D accuracy remains near 51\% across the earliest boundaries, close to the performance obtained by restoring all MLP layers. As the boundary moves deeper, dynamic accuracy declines substantially and eventually approaches the specialized model's performance. Meanwhile, VSI-Bench accuracy rises sharply at later boundaries, indicating that retaining more specialized MLP layers favors spatial reasoning at the expense of dynamic recovery.

We select $b=4$ within the early plateau: restoring layers $j>4$ recovers nearly the same dynamic performance as restoring all MLPs, with little change in spatial accuracy. This suggests that reverting the first four layers provides limited additional benefit for dynamic recovery. 
Accordingly, we define MLP layers \(j>4\) as the interference-sensitive group in D-EMA, allowing stronger consolidation to be assigned to these layers when differentiated averaging horizons are used.

\subsection{Weight interpolation}
\label{sec:weight_interpolation}

\subsubsection{Interpolation on GeoThinker}

\begin{table}[h!]
  \centering
  \caption{
  \textbf{Parameter-group interpolation between GeoThinker and the pretrained Qwen3-VL-8B model.}We interpolate the specialized checkpoint $\theta_T$ with the pretrained checkpoint $\theta_0$, either globally or within selected parameter groups. The interpolation ratio denotes the contribution of $\theta_T$ in the selected group. Remaining parameters retain their specialized values. Early and late MLP groups correspond to layer indices $j\leq4$ and $j>4$, respectively. Scores outside parentheses use deterministic rule-based answer parsing, while parenthesized values report direct exact-match accuracy without parsing.
  }
  \label{tab:appendix_qwen3vl8b_interpolate}
  \setlength{\tabcolsep}{3pt}
  \begin{tabular}{l | c |c c}
    \toprule
    \multirow{2}{*}{\textbf{Model}} & \textbf{Interpolate} & \multirow{2}{*}{\textbf{VSI}} & \multirow{2}{*}{\textbf{VLM4D}}\\
     & \textbf{ratio} & & \\
    \midrule
    \rowcolor{black!11}
    \multicolumn{4}{l}{\textbf{Baseline}} \\

    GeoThinker\textsubscript{Qwen3-VL-8B} ($\theta_{T}$) & 100\% & 72.6 & 42.40 \\
    Qwen3-VL-8B-Instruct ($\theta_{0}$) & 0\% & 57.75 & 60.07 (60.07) \\

    \midrule
    \rowcolor{black!11}
    \multicolumn{4}{l}{\textbf{Interpolate all}} \\

    75\% $\theta_{T}$ + 25\% $\theta_{0}$ & 75\% & 71.70 & 45.95 \\
    50\% $\theta_{T}$ + 50\% $\theta_{0}$ & 50\% & 70.12 & 49.17 \\
    25\% $\theta_{T}$ + 75\% $\theta_{0}$ & 25\% & 64.34 & 53.68 \\
    12.5\% $\theta_{T}$ + 87.5\% $\theta_{0}$ & 12.5\% & 59.36 & 54.62 \\

    \midrule
    \rowcolor{black!11}
    \multicolumn{4}{l}{\textbf{Interpolate all MLP}} \\

    100\%$\theta_{T,Attn}$ + 25\%$\theta_{0,MLP}$ + 75\%$\theta_{T,MLP}$ & 75\% & 71.86 & 44.98 \\
    100\%$\theta_{T,Attn}$ + 50\%$\theta_{0,MLP}$ + 50\%$\theta_{T,MLP}$ & 50\% & 70.72 & 46.58 \\
    100\%$\theta_{T,Attn}$ + 75\%$\theta_{0,MLP}$ + 25\%$\theta_{T,MLP}$ & 25\% & 67.46 & 49.39 \\
    100\%$\theta_{T,Attn}$ + 87.5\%$\theta_{0,MLP}$ + 12.5\%$\theta_{T,MLP}$ & 12.5\% & 64.80 & 50.88 \\
    100\%$\theta_{T,Attn}$ + 100\%$\theta_{0,MLP}$ & 0\% & 61.43 & 51.59 \\

    \midrule
    \rowcolor{black!11}
    \multicolumn{4}{l}{\textbf{Interpolate all Attn}} \\

    100\%$\theta_{T,MLP}$ + 25\%$\theta_{0,Attn}$ + 75\%$\theta_{T,Attn}$ & 75\% & 72.06 & 43.11 \\
    100\%$\theta_{T,MLP}$ + 50\%$\theta_{0,Attn}$ + 50\%$\theta_{T,Attn}$ & 50\% & 71.53 & 43.99 \\
    100\%$\theta_{T,MLP}$ + 75\%$\theta_{0,Attn}$ + 25\%$\theta_{T,Attn}$ & 25\% & 70.72 & 45.15 \\
    100\%$\theta_{T,MLP}$ + 87.5\%$\theta_{0,Attn}$ + 12.5\%$\theta_{T,Attn}$ & 12.5\% & 70.03 & 45.98 \\
    100\%$\theta_{T,MLP}$ + 100\%$\theta_{0,Attn}$ & 0\% & 69.51 & 46.86 \\

    \midrule
    \rowcolor{black!11}
    \multicolumn{4}{l}{\textbf{Interpolate $MLP_{\le 4}$}} \\

    100\%$\theta_{T,\{Attn,MLP>4\}}$ + 25\%$\theta_{0,MLP\le4}$ + 75\%$\theta_{T,MLP\le4}$ & 75\% & 72.09 & 42.67 \\
    100\%$\theta_{T,\{Attn,MLP>4\}}$ + 50\%$\theta_{0,MLP\le4}$ + 50\%$\theta_{T,MLP\le4}$ & 50\% & 72.19 & 43.00 \\
    100\%$\theta_{T,\{Attn,MLP>4\}}$ + 75\%$\theta_{0,MLP\le4}$ + 25\%$\theta_{T,MLP\le4}$ & 25\% & 72.25 & 43.22 \\
    100\%$\theta_{T,\{Attn,MLP>4\}}$ + 87.5\%$\theta_{0,MLP\le4}$ + 12.5\%$\theta_{T,MLP\le4}$ & 12.5\% & 72.31 & 43.17 \\
    100\%$\theta_{T,\{Attn,MLP>4\}}$ + 100\%$\theta_{0,MLP\le4}$ & 0\% & 72.48 & 43.61 \\

    \midrule
    \rowcolor{black!11}
    \multicolumn{4}{l}{\textbf{Interpolate $MLP_{>4}$}} \\

    100\%$\theta_{T,\{Attn,MLP\le4\}}$ + 25\%$\theta_{0,MLP>4}$ + 75\%$\theta_{T,MLP>4}$ & 75\% & 71.70 & 44.32 \\
    100\%$\theta_{T,\{Attn,MLP\le4\}}$ + 50\%$\theta_{0,MLP>4}$ + 50\%$\theta_{T,MLP>4}$ & 50\% & 70.74 & 46.53 \\
    100\%$\theta_{T,\{Attn,MLP\le4\}}$ + 75\%$\theta_{0,MLP>4}$ + 25\%$\theta_{T,MLP>4}$ & 25\% & 67.12 & 49.33 \\
    100\%$\theta_{T,\{Attn,MLP\le4\}}$ + 87.5\%$\theta_{0,MLP>4}$ + 12.5\%$\theta_{T,MLP>4}$ & 12.5\% & 64.32 & 50.44 \\
    100\%$\theta_{T,\{Attn,MLP\le4\}}$ + 100\%$\theta_{0,MLP>4}$ & 0\% & 61.06 & 51.15 \\

    \bottomrule
  \end{tabular}
\end{table}

Table~\ref{tab:appendix_qwen3vl8b_interpolate} examines which pretrained parameter groups are most sensitive to spatial specialization. Global interpolation reveals an acquisition--retention trade-off: retaining 50\% of the specialized weights recovers VLM4D accuracy from 42.40 to 49.17 while maintaining a VSI-Bench score of 70.12.

Group-wise interpolation further localizes the recoverable interference. Restoring all MLP parameters to their pretrained values raises VLM4D accuracy to 51.59, compared with 46.86 when restoring attention parameters alone. Within the MLP blocks, restoring layers $j\leq4$ yields only a small improvement to 43.61, whereas restoring layers $j>4$ reaches 51.15, nearly matching the recovery obtained by restoring all MLPs. These results identify later MLP blocks as a key interference-sensitive group, supporting the boundary $b=4$ and stronger temporal consolidation for these parameters in D-EMA.

\subsubsection{Interpolation on \name without D-EMA}

\begin{table}[h!]
  \centering
  \caption{\textbf{Parameter-group interpolation for DynThinker without D-EMA.}
We interpolate the spatially specialized Qwen3-VL-4B checkpoint $\theta_T$ with its pretrained counterpart $\theta_0$, globally or within selected parameter groups. The interpolation ratio denotes the contribution of $\theta_T$ in the interpolated group. All remaining parameters retain their specialized values. Early and late MLP groups correspond to layer indices $j\leq4$ and $j>4$, respectively. Scores outside parentheses use deterministic rule-based answer parsing, while parenthesized values report direct exact-match accuracy without parsing.
}
  \label{tab:appendix_qwen3vl4b_interpolate}
  \setlength{\tabcolsep}{3pt}
  \begin{tabular}{l | c |c c}
    \toprule
    \multirow{2}{*}{\textbf{Model}} & \textbf{Interpolate} & \multirow{2}{*}{\textbf{VSI}} & \multirow{2}{*}{\textbf{VLM4D}}\\
     & \textbf{ratio} & & \\
    \midrule
    \rowcolor{black!11}
    \multicolumn{4}{l}{\textbf{Baseline}} \\

    \name\textsubscript{Qwen3-VL-4B} w/o D-EMA ($\theta_{T}$) & 100\% & 71.06 & 50.33 \\
    Qwen3-VL-4B-Instruct ($\theta_{0}$) & 0\% & 54.90 & 57.10 (1.87) \\

    \midrule
    \rowcolor{black!11}
    \multicolumn{4}{l}{\textbf{Interpolate all}} \\

    75\% $\theta_{T}$ + 25\% $\theta_{0}$ & 75\% & 71.32 & 48.84 \\
    50\% $\theta_{T}$ + 50\% $\theta_{0}$ & 50\% & 71.01 & 45.26 \\
    25\% $\theta_{T}$ + 75\% $\theta_{0}$ & 25\% & 66.93 & 58.09 \\
    12.5\% $\theta_{T}$ + 87.5\% $\theta_{0}$ & 12.5\% & 61.84 & 61.45 \\

    \midrule
    \rowcolor{black!11}
    \multicolumn{4}{l}{\textbf{Interpolate all MLP}} \\

    100\%$\theta_{T,Attn}$ + 25\%$\theta_{0,MLP}$ + 75\%$\theta_{T,MLP}$ & 75\% & 71.44 & 52.32 \\
    100\%$\theta_{T,Attn}$ + 50\%$\theta_{0,MLP}$ + 50\%$\theta_{T,MLP}$ & 50\% & 71.08  & 53.35 \\
    
    100\%$\theta_{T,Attn}$ + 75\%$\theta_{0,MLP}$ + 25\%$\theta_{T,MLP}$ & 25\% & 70.07 & 55.83\\
    
    100\%$\theta_{T,Attn}$ + 87.5\%$\theta_{0,MLP}$ + 12.5\%$\theta_{T,MLP}$ & 12.5\% & 67.81 & 57.59 \\
    
    100\%$\theta_{T,Attn}$ + 100\%$\theta_{0,MLP}$ & 0\% & 64.57 & 58.03\\

    \midrule
    \rowcolor{black!11}
    \multicolumn{4}{l}{\textbf{Interpolate all Attn}} \\

    100\%$\theta_{T,MLP}$ + 25\%$\theta_{0,Attn}$ + 75\%$\theta_{T,Attn}$ & 75\% & 71.17 & 50.44 \\
    100\%$\theta_{T,MLP}$ + 50\%$\theta_{0,Attn}$ + 50\%$\theta_{T,Attn}$ & 50\% & 70.50 & 49.77 \\
    
    100\%$\theta_{T,MLP}$ + 75\%$\theta_{0,Attn}$ + 25\%$\theta_{T,Attn}$ & 25\% & 70.29 & 48.84\\
    
    100\%$\theta_{T,MLP}$ + 87.5\%$\theta_{0,Attn}$ + 12.5\%$\theta_{T,Attn}$ & 12.5\% & 69.91 & 48.07 \\
    
    100\%$\theta_{T,MLP}$ + 100\%$\theta_{0,Attn}$ & 0\% & 68.90 & 47.79\\

    \midrule
    \rowcolor{black!11}
    \multicolumn{4}{l}{\textbf{Interpolate $MLP_{\le 4}$}} \\

    100\%$\theta_{T,\{Attn,MLP>4\}}$ + 25\%$\theta_{0,MLP\le4}$ + 75\%$\theta_{T,MLP\le4}$ & 75\% & 71.07 & 50.55 \\
    
    100\%$\theta_{T,\{Attn,MLP>4\}}$ + 50\%$\theta_{0,MLP\le4}$ + 50\%$\theta_{T,MLP\le4}$ & 50\% & 70.94 & 50.71 \\
    
    100\%$\theta_{T,\{Attn,MLP>4\}}$ + 75\%$\theta_{0,MLP\le4}$ + 25\%$\theta_{T,MLP\le4}$ & 25\% & 71.12 & 50.77\\
    
    100\%$\theta_{T,\{Attn,MLP>4\}}$ + 87.5\%$\theta_{0,MLP\le4}$ + 12.5\%$\theta_{T,MLP\le4}$ & 12.5\% & 70.88 & 51.10 \\
    
    100\%$\theta_{T,\{Attn,MLP>4\}}$ + 100\%$\theta_{0,MLP\le4}$ & 0\% & 70.81 & 50.77 \\

    \midrule
    \rowcolor{black!11}
    \multicolumn{4}{l}{\textbf{Interpolate $MLP_{>4}$}} \\

    100\%$\theta_{T,\{Attn,MLP\le4\}}$ + 25\%$\theta_{0,MLP>4}$ + 75\%$\theta_{T,MLP>4}$ & 75\% & 71.15 & 51.81 \\
    
    100\%$\theta_{T,\{Attn,MLP\le4\}}$ + 50\%$\theta_{0,MLP>4}$ + 50\%$\theta_{T,MLP>4}$ & 50\% & 70.89 & 53.13 \\
    
    100\%$\theta_{T,\{Attn,MLP\le4\}}$ + 75\%$\theta_{0,MLP>4}$ + 25\%$\theta_{T,MLP>4}$ & 25\% & 69.55 & 54.46 \\
    
    100\%$\theta_{T,\{Attn,MLP\le4\}}$ + 87.5\%$\theta_{0,MLP>4}$ + 12.5\%$\theta_{T,MLP>4}$ & 12.5\% & 67.05 & 56.55 \\
    
    100\%$\theta_{T,\{Attn,MLP\le4\}}$ + 100\%$\theta_{0,MLP>4}$ & 0\% & 63.51 & 58.31\\

    \bottomrule
  \end{tabular}
\end{table}

Table~\ref{tab:appendix_qwen3vl4b_interpolate} extends our interpolation analysis to \name trained without D-EMA. 3D-focused post-training improves VSI-Bench accuracy from 54.90\% to 71.06\%, while reducing VLM4D accuracy from 57.10\% to 50.33\%. Thus, integrating spatial and dynamic geometry features alone does not prevent spatial--dynamic interference during adaptation.

The recovery of dynamic performance depends strongly on the interpolated parameter group. 
Notably, fully restoring the late-layer MLPs raises VLM4D to 58.31\%, exceeding both the specialized checkpoint at 50.33\% and the pretrained backbone at 57.10\%. This suggests that capability recovery does not require reverting the entire model. Instead, recombining preserved late-layer representations with adapted components can yield complementary benefits.
In contrast, restoring early-layer MLPs yields only 50.77\%, while restoring attention parameters reduces accuracy to 47.79\%. 
Partial interpolation also offers a favorable trade-off: retaining 50\% of the specialized late-layer MLP weights improves VLM4D accuracy to 53.13\%, with VSI-Bench accuracy remaining close to the specialized checkpoint (70.89\% vs.\ 71.06\%). These findings corroborate the sensitivity of late-layer MLPs and support their selective consolidation in D-EMA with boundary $b=4$.

\section{Additional results (downstream)}
\subsection{Embodied Manipulation}
\label{sec:appendix_embodied_manipulation}

\subsubsection{Setting} 
\textbf{Baseline.} StarVLA~\citep{community2026starvla} is a Lego-like, open-source framework for developing vision-language-action (VLA) models with modular backbones and action heads. We adopt StarVLA-OFT as our primary embodied baseline, which provides a simple and effective VLA architecture using an MLP head to decode continuous actions.

\textbf{Evaluation protocol:} We build the embodied variant of \name upon StarVLA-OFT by incorporating our proposed SDF into its VLM backbone to selectively integrate spatial-dynamic priors for action prediction. Following the StarVLA-OFT setup on LIBERO~\citep{liu2023libero}, we train for 30,000 steps with a batch size of 16 and a learning rate of $2.5\times10^{-5}$.

\subsubsection{Evaluation Results.} 

\begin{table}[h!]
  \centering
  \caption{\textbf{Comparison of different VLA models on LIBERO.} 
  We train one policy for all 4 suites. All scores are averaged over 500 trials for each task suite (10 tasks × 50 episodes). %
  }
  \label{tab:appendix_libero_results}
  \setlength{\tabcolsep}{3pt}
  \begin{tabular}{l | c c c c c | c}
    \toprule
    \textbf{Model} & \textbf{Steps} & \textbf{Spatial} & \textbf{Object} & \textbf{Goal} & \textbf{Long} & \textbf{Avg} \\
    \midrule
    \rowcolor{black!11}
    \multicolumn{7}{l}{\textbf{Baseline}} \\
    $\pi_0$+FAST~\citep{pertsch2025fast} & - & 96.4 & 96.8 & 88.6 & 60.2 & 85.5 \\
    $\pi_0$~\citep{black2024pi0} & - & 96.8 & 98.8 & 95.8 & 85.2 & 94.1 \\
    OpenVLA-OFT-7B~\citep{kim2025fine} & 175K & 97.6 & 98.4 & 97.9 & 94.5 & \sota{97.1} \\
    GR00T-N1.5~\citep{bjorck2025gr00t} & 20K  & 92.0 & 92.0 & 86.0 & 76.0 & 86.5 \\

    \midrule
    \rowcolor{black!11}
    \multicolumn{7}{l}{\textbf{Qwen3-VL-4B}} \\

    StarVLA-OFT~\citep{community2026starvla}  & 30K & 97.8 & 98.6 & 96.2 & 93.8 & 96.6 \\

    GeoThinker~\citep{li2026thinking}  & 30K & 98.4 & 99.0 & 98.4 & 91.2 & 96.7 \\
    \rowcolor{navyblue!10} 
    \name\textit{(Ours)} & 30K & 99.0 & 99.4 & 98.0 & 92.8 & \sota{97.3} \\

    \bottomrule
  \end{tabular}
\end{table}

As shown in Table~\ref{tab:appendix_libero_results}, \name achieves the highest observed average success rate of 97.3\% with a single policy across all four LIBERO suites. Under the same Qwen3-VL-4B backbone and 30K-step training schedule, it improves over StarVLA-OFT and GeoThinker by 0.7 and 0.6 percentage points, respectively. \name achieves the highest success rates on the Spatial and Object suites and improves the Long-suite result of GeoThinker from 91.2\% to 92.8\%. These results indicate that SDF transfers effectively to language-conditioned robot manipulation.

\end{document}